\documentclass[runningheads]{llncs}

\usepackage[mobile]{eccv}

\usepackage{eccvabbrv}

\usepackage{graphicx}
\usepackage{booktabs}

\usepackage[accsupp]{axessibility}  % Improves PDF readability for those with disabilities.

\usepackage{hyperref}

\usepackage{orcidlink}
\usepackage{multirow} 
\usepackage{algorithm}
\usepackage{algpseudocode}

\newcommand{\parsection}[1]{\noindent\textbf{#1} }

\begin{document}

\title{Gaussian Grouping: Segment and Edit Anything in 3D Scenes} 



% \author{Mingqiao Ye\orcidlink{0009-0009-9368-2197} \and
% Martin Danelljan\orcidlink{0000-0001-6144-9520} \and
% Fisher Yu\orcidlink{0000-0001-8829-7344} \and
% Lei Ke\orcidlink{0000-0002-0944-720X}}

\author{
 Mingqiao Ye \and Martin Danelljan \and Fisher Yu \and Lei Ke\small{$~^{\spadesuit}$} \\
 }

\authorrunning{M.~Ye et al.}

% Springer Heidelberg, Tiergartenstr.~17, 69121 Heidelberg, Germany

\institute{Computer Vision Lab, ETH Zurich}

\maketitle

 % \footnote{\small{$^{\spadesuit}$~Project Lead}}

\renewcommand{\thefootnote}{\fnsymbol{footnote}}
\footnotetext{\small{$^{\spadesuit}$~Project Lead}}

\begin{center} 
\vspace{-0.2in}
\centering 
\includegraphics[width=1.0\linewidth]{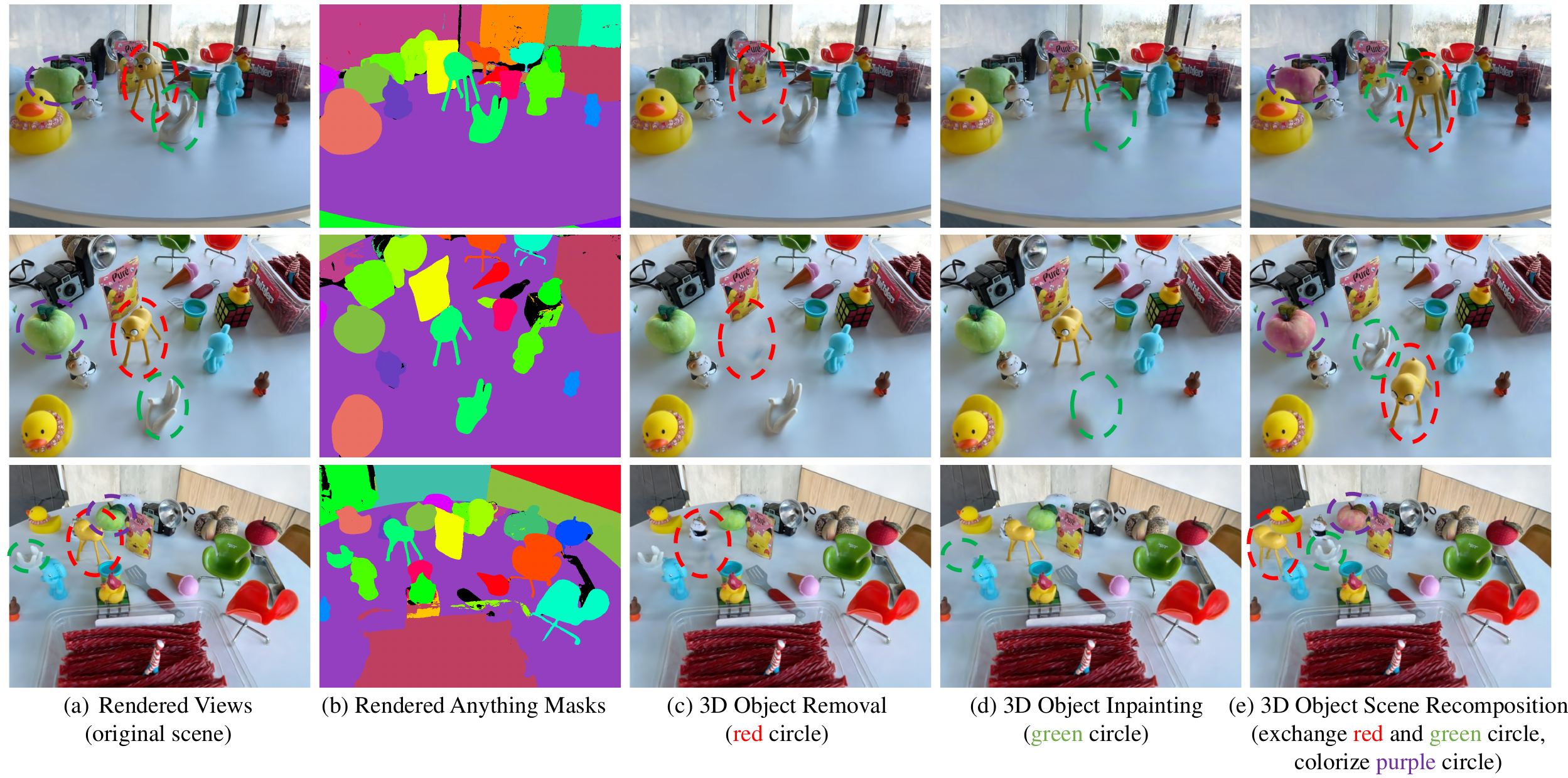}
	\vspace{-0.3in}
	\captionof{figure}{Our Gaussian Grouping jointly reconstructs (column a) and segments (column b) anything in full open-world 3D scenes, with fine-grained instance and stuff level modeling. This enables versatile scene editing applications, such as 3D object removal (column c), 3D object inpainting (column d, which first removes the 3D object and then inpaints the holes) and scene re-composition and object colorization (column e). Since the segmentation information is encapsulated in the 3D Gaussians, editing tasks such as 3D object removal, colorization and object location exchange can be performed directly \textbf{w/o} training, while inpainting only requires minutes of fine-tuning.}
	\label{fig:teaser}
\vspace{-3mm}
\end{center}

\begin{abstract}
The recent Gaussian Splatting achieves high-quality and real-time novel-view synthesis of the 3D scenes.
However, it is solely concentrated on the appearance and geometry modeling, while lacking in fine-grained object-level scene understanding.
To address this issue, we propose Gaussian Grouping, which extends Gaussian Splatting to jointly reconstruct and segment anything in open-world 3D scenes.
We augment each Gaussian with a compact Identity Encoding, allowing the Gaussians to be grouped according to their object instance or stuff membership in the 3D scene.
Instead of resorting to expensive 3D labels, we supervise the Identity Encodings during the differentiable rendering by leveraging the 2D mask predictions by Segment Anything Model (SAM), along with introduced 3D spatial consistency regularization.
Compared to the implicit NeRF representation, we show that the discrete and grouped 3D Gaussians can reconstruct, segment and edit anything in 3D with high visual quality, fine granularity and efficiency.
Based on Gaussian Grouping, we further propose a local Gaussian Editing scheme, which shows efficacy in versatile scene editing applications, including 3D object removal, inpainting, colorization, style transfer and scene recomposition.
Our code and models are at \href{https://github.com/lkeab/gaussian-grouping}{\textcolor{red}{github.com/lkeab/gaussian-grouping}}.
\end{abstract}

\section{Introduction}
\label{sec:intro}

Open-world 3D scene understanding is an essential challenge, with far-reaching implications for robotics, AR / VR, and autonomous driving. 
Given a set of posed RGB images, our goal is to learn an effective 3D representation that jointly reconstructs and segments anything in the 3D scene. 
The representation should easily support a wide range of downstream scene editing applications.
For example, in Figure~\ref{fig:teaser}, the 3D object of the scene can be easily removed or inpainted, and the scene can be recomposed by exchanging object locations.

While there has been remarkable progress in 2D scene understanding brought by SAM and its variants~\cite{kirillov2023segment,sam_hq,mobile_sam}, their extension to 3D has been constrained.
This is mostly due to the availability and the labor-intensive process of creating 3D scene datasets.
Existing methods~\cite{Schult23,dai2017scannet} rely on manually-labeled datasets, which are both costly and limited in scope, or require accurately scanned point clouds~\cite{takmaz2023openmask3d,peng2023openscene} as input.
This hinders the development of the 3D scene understanding methods that can quickly generalize across various real-world scenes.

By taking multi-view captures, existing NeRF-based methods~\cite{siddiqui2023panoptic,liu2023instance,lerf2023,kobayashi2022decomposing} lift 2D masks or distill CLIP~\cite{radford2021learning} / DINO~\cite{caron2021emerging} features via neural fields rendering. 
However, due to the implicit and continuous representation of NeRF, these methods require expensive random sampling and are computationally intensive to optimize.
Further, it is hard to directly adjust NeRF-based approaches for the downstream local editing tasks~\cite{kobayashi2022decomposing}, because the learned neural networks, such as MLPs, cannot decompose each part or module in the 3D scene easily.
Several methods~\cite{shen2023anything,cen2023segment} combine NeRF or stable-diffusion~\cite{rombach2022high} with the masks of SAM, but they only focus on a single object. 

Alternative to NeRFs, the recently emerged 3D Gaussian Splatting~\cite{kerbl20233d} has shown impressive reconstruction quality with high training and rendering efficiency. It represents the 3D scene with an array of colored and explicit 3D Gaussians, which are rendered into camera views for novel view synthesis.
Nevertheless, Gaussian Splatting does not model object instances or semantic understanding.
To achieve fine-grained scene understanding, we extend this approach beyond merely capturing the scene's appearance and geometry to include the individual objects and elements that constitute the 3D environments.

We propose~\textit{Gaussian Grouping}, which 
represents the whole 3D scene with a set of grouped 3D Gaussians.
By inputting multi-view captures and the corresponding automatically generated masks by SAM, our method learns a discrete and grouped 3D representation for reconstructing and segmenting anything in the 3D scene.
Gaussian Grouping inherits SAM's strong zero-shot 2D scene understanding capability and extends it to the 3D space by producing consistent novel view synthesis and segmentation.

To this end, instead of solely modeling the scene appearance and geometry, our approach also captures the identity of each Gaussian of the 3D scene by grouping. 
A new property, our \textit{Identity Encoding}, is augmented to each Gaussian.
The Identity Encoding is a compact and low-dimensional learnable embedding.
To leverage the segmentation supervision in 2D, Identity Encoding is trained via differentiable Gaussian rendering, where the encoding vectors of various Gaussians are splatted onto the 2D rendering view. Then, we take the 2D rendered identity features and employ an extra linear layer to classify these splatted embeddings on each 2D location for identity classification. 

To further boost the grouping accuracy, besides the standard cross-entropy loss for identity classification, we also introduce an \textbf{un}-supervised 3D Regularization Loss by leveraging 3D spatial consistency.
The loss enforces the Identity Encodings of the top $K$-nearest 3D Gaussians to be close in their feature distance. We find it helps Gaussians inside the 3D object or heavily occluded to be supervised during training more sufficiently. 

We highlight the advantages of Gaussian Grouping not only by providing high visual quality and fast training, but also through its downstream scene editing applications brought by our dense 3D segmentation.
Since our model captures the 3D scenes as compositional structures, each group of 3D Gaussians operates independently, allowing for parts to be fully decoupled or separated. 
This decomposition is crucial in scenarios where individual components need to be identified, manipulated, or replaced \textbf{w/o} disrupting the entire scene structure.

\textit{Our contribution can be summarized as follows}:
\vspace{-0.1in}
\begin{enumerate}
    \item [1.] We introduce Gaussian Grouping, the first 3D Gaussian Splatting-based segmentation framework that lifts knowledge of SAM to 3D scene anything zero-shot segmentation without the need for 3D mask labels.
    \item [2.] Gaussian Grouping supports various downstream tasks with our proposed Local Gaussian Editing scheme. Individual components are identified, manipulated, or replaced without disrupting the entire scene structure. By the grouped 3D Gaussians, we show extensive 3D scene editing cases, such as 3D scene re-composition, object inpainting, object removal and object style transfer, with both impressive visual effect and fine granularity.
    \item [3.] Our training and rendering processes are swift, ensuring compliance with real-time operational requirements.
\end{enumerate}

\vspace{-0.2in}

\section{Related Works}

\subsubsection{3D Gaussian Models}
3D Gaussian Splatting~\cite{kerbl20233d} has recently emerged as a powerful approach to reconstruct 3D scenes via real-time radiance field rendering. 
Several follow-up methods~\cite{luiten2023dynamic,yang2023deformable,yang2023real} extend it to dynamic 3D scenes by tracking of all dense scene elements~\cite{luiten2023dynamic} or deformation field modeling~\cite{yang2023deformable,wu20234d}.
Another stream of works~\cite{chen2023text,tang2023dreamgaussian,yi2023gaussiandreamer} focuses on the 3D content creation, which combines Gaussian Splatting with diffusion-based models and shows high-quality generation results.
However, none of the existing Gaussian Splatting works enables object / stuff-level or semantic understanding of the 3D scene.
Our Gaussian Grouping extends Gaussian Splatting from pure scene appearance and geometry modeling to also support open-world and fine-grained scene understanding.
We show that the grouped 3D Gaussians is an effective and flexible 3D representation in supporting a series of downstream scene editing applications. 
\vspace{-0.1in}

\subsubsection{Radiance-based Open World Scene Understanding}
Existing semantic scene modeling approaches combined with radiance fields are NeRF-based~\cite{mildenhall2020nerf,mazur2023feature,zhang2023nerflets}.
Semantic-NeRF~\cite{Zhi:etal:ICCV2021} initiated the incorporation of semantic information into NeRFs and facilitated the generation of semantic masks from novel views. 
Building upon it, subsequent research focusing on in-domain scene modeling, has expanded the scope by introducing instance modeling~\cite{kundu2022panoptic,fu2022panoptic,wang2022dm,siddiqui2023panoptic} or encoding visual features~\cite{tschernezki2022neural, kobayashi2022decomposing,tschernezki22neural,vora2021nesf,rebain2021derf} that support semantic delineation. Unlike our approach, most of these methods are designed for in-domain scene modeling and cannot generalize to open-world scenarios.

By distilling 2D features such as CLIP~\cite{radford2021learning} or DINO~\cite{caron2021emerging}, Distilled Feature Fields~\cite{kobayashi2022decomposing} explores embedding pixel-aligned feature vectors, and LERF~\cite{lerf2023} learns language field to help open-world 3D semantics.
In contrast to our work, these open-world approaches are only limited in semantic segmentation (hard to separate similar objects of the same category) and cannot produce very accurate segmentation masks as shown in our experiment. To our knowledge, we propose the first Gaussian-based method to tackle open-world 3D scene understanding, where we show the advantages compared to existing NeRF-based approaches~\cite{lerf2023,kobayashi2022decomposing,siddiqui2023panoptic} in segmentation quality, efficiency and good extension to downstream scene editing applications.
\vspace{-0.1in}

\subsubsection{SAM in 3D}
Segment Anything Model (SAM)~\cite{kirillov2023segment} was released as a foundational vision model for zero-shot 2D segmentation. Several works lift SAM's 2D masks to 3D segmentation via NeRF~\cite{cen2023segment,chen2023interactive} or 3D point cloud~\cite{yang2023sam3d}. However, these NeRF-based approaches only focus on a single / few objects of the 3D scene, while our Gaussian Grouping operates in automatic \textit{everything} mode to attain the holistic understanding of each instance / stuff of the full scene.
\vspace{-0.1in}

\subsubsection{Radiance-based Scene Editing}
Manipulating 3D scenes via a radiance field is challenging. For existing NeRF-based scene editing / manipulation works~\cite{yuan2022nerf,liu2021editing,wu2022object,kania2022conerf,liu2022nerf}, 
Clip-NeRF~\cite{clipnerf}, Object-NeRF~\cite{yang2021learning}, and LaTeRF~\cite{laterf} design approaches to change and complete objects represented by NeRFs; however, their application scenario is limited to simple objects, rather than scenes with significant clutter and texture. 
Some other works specify bounding boxes~\cite{ost2021neural,orf}, to allow flexible compositing of various objects~\cite{zhang2021editable} or combining with physical simulation~\cite{Li2023ClimateNeRF}.
Most recently, 3D object inpainting is studied in SPIn-NeRF~\cite{mirzaei2023spin} and language-based scene editing is proposed in~\cite{instructnerf2023}.
Compared to them, we apply our Gaussian Grouping to versatile scene editing tasks, and show benefits brought by the fine-grained scene modeling, where multiple scene editings can be easily superimposed on the same image. The design of Local Gaussian Editing makes the whole training / editing process both simple and efficient.

\section{Method}
Our work aims to build an expressive 3D scene representation, which not only models appearance and geometry, but also captures every instance and stuff identity of the scene.
We design our method based on the recent 3D Gaussian Splatting~\cite{kerbl20233d}, and extend it from pure 3D reconstruction to fine-grained scene understanding. 
Our approach, called \textit{Gaussian Grouping}, is capable of: 
1) modeling each 3D part of the scene with appearance, geometry together with their mask identities; 
2) fully decomposing the 3D scene into discrete \emph{groups}, \eg representing different object instances to enable editing; 3) allowing for fast training and rendering, while not diluting the original 3D reconstruction quality.

Gaussian Grouping effectively leverages the dense 2D mask proposals of SAM, and lifts them to segment anything in the 3D scene via radiance fields rendering.
In Sec~\ref{sec:gau_splat}, we first briefly review the 3D Gaussian Splatting method on radiance field rendering.
We then detail the input data pre-processing steps and further describe the proposed Gaussian Grouping in Section~\ref{sec:gau_group}.
With the constructed 3D representation, finally, we show its advantages in downstream scene editing tasks by the efficient Local Gaussian Editing in Section~\ref{sec:partial_tuning}.

% an overview figure of the method

\subsection{Preliminaries: 3D Gaussian Splatting}
\label{sec:gau_splat}

Gaussian Splatting, as introduced by \cite{kerbl20233d}, encapsulates 3D scene information using a suite of 3D colored Gaussians. This technique has established its effectiveness in the reconstruction tasks, exhibiting high inference speeds and remarkable quality of reconstruction within timeframes on par with those of NeRF. Yet, its potential in scene understanding has not been thoroughly investigated. Our research reveals that 3D Gaussians hold considerable promise for the open-world and complex 3D scene understanding.

To represent the scene, each Gaussian's property is characterized by a centroid \(\mathbf{p} = \{x, y, z\} \in \mathbb{R}^3\), a 3D size \(\mathbf{s} \in \mathbb{R}^3\) in standard deviations, and a rotational quaternion \(\mathbf{q} \in \mathbb{R}^4\). To allow fast $\alpha$-blending for rendering, an opacity value \(\alpha \in \mathbb{R}\) and a color vector \(\mathbf{c}\) are represented in the three degrees of spherical harmonics (SH) coefficients. These adjustable parameters are collectively symbolized by \(S_{\Theta_i}\), where \(S_{\Theta_i} = \{\mathbf{p}_i, \mathbf{s}_i, \mathbf{q}_i, \alpha_i, \mathbf{c}_i\}\) represents the set of parameters for the \(i\)-th Gaussian.
Gaussian Splatting projects these 3D Gaussians onto the 2D image plane and implement differentiable rendering for each pixel.

\subsection{3D Gaussian Grouping}
\label{sec:gau_group}
In this section, we describe the design of our Gaussian Grouping. To enable 3D Gaussians for fine-grained scene understanding, our key insight is that we preserve all attributes of the Gaussians (such as their position, color, opacity, and size) in their original setting, but add new Identity Encoding parameters (similar to the format of color modeling). This allows each Gaussian to be assigned to its represented instances or stuff in the 3D scene.

\begin{figure*}[!t]
	\centering
	\vspace{-0.0in}
	\includegraphics[width=1.0\linewidth]{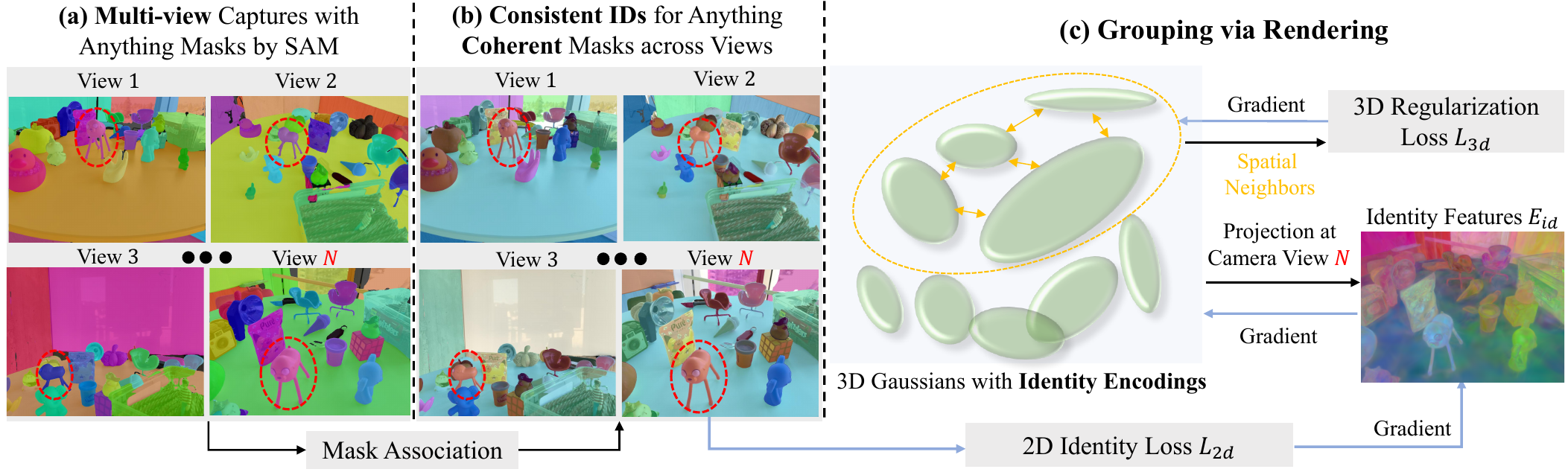}
	\vspace{-0.25in}
	\caption{The method pipeline of our Gaussian Grouping contains three main steps: \textbf{(a)} We first prepare the input by deploying SAM to automatically generate masks in \textit{everything} mode for each view independently.~\textbf{(b)} Then, to obtain the consistent mask IDs across training views, we take a universal temporal propagation model~\cite{cheng2023tracking} to associate the mask labels and generate a coherent multi-view segmentation.~\textbf{(c)} With the prepared training input, we jointly learn all properties of the 3D Gaussians, including their group Identity Encoding, by differentiable rendering. Our encoding is supervised by the 2D Identity Loss, leveraging the coherent segmentation views, and a 3D Regularization loss. We use color the denote object IDs across frames for input views. We omit the rendering process for other Gaussian parameters and the density control part for simplicity, as it is inherited from~\cite{kerbl20233d}.}
	\label{fig:model}
	\vspace{-0.25in}
\end{figure*}

% \begin{algorithm*}[!t]
\begin{algorithm}[H]
    % \footnotesize % or \scriptsize for even smaller text
    \scriptsize
    \caption{\textit{Gaussian Grouping}
    }
		\label{alg:optimization}
		\begin{algorithmic}
			\State $p \gets$ SfM Points	\Comment{3D Positions}
                \State $m = (m_1, m_2, \ldots, m_K ) \gets$ SAM	\Comment{\textbf{SAM's Masks} at Various $K$ Views}
                \State $(\hat{M}_1, \hat{M}_2, \ldots, \hat{M}_K ) \gets$ Zero-shot Tracking(m)	\Comment{Multi-view \textbf{Associated Masks}}
			\State $s, \alpha, c, \textbf{e} \gets$ InitAttributes() \Comment{Covariances, Opacities,  Colors, \textbf{Identity Encodings}}
			\State $i \gets 0$	\Comment{Iteration Count}
			
			\While{not converged}
			
			\State $V, \hat{I}, \hat{M} \gets$ SampleTrainingView()	\Comment{Camera View $V$, Image and \textbf{Mask}}
			\State $I, \textbf{E}_\text{id} \gets$ Rasterize($p$, $s$, $a$, $c$, $\textbf{e}$ ,$V$)	\Comment{Rendered Image and \textbf{Identity Encoding}}
			
			\State $\mathcal{L}_\text{image} \gets \mathcal{L}(I, \hat{I}) $ \Comment{Original Image Rendering Loss}
			\textcolor{red}{\State $L_\text{id} \gets \lambda_{2d}\mathcal{L}_{\text{2d}}(\textbf{E}_\text{id}, \hat{M}) + \lambda_{3d}\mathcal{L}_{\text{3d}}(\textbf{e}) $} \Comment{\textbf{Identity Grouping Loss}, Eq.~\ref{eq:reg_3d}}
                \textcolor{red}{\State $\mathcal{L} \gets$ $\mathcal{L}_\text{image} + \mathcal{L}_\text{id} $} \Comment{Total Loss}
			\State $p$, $s$, $a$, $c$, $\textbf{e}$ $\gets$ Adam($\nabla \mathcal{L}$) \Comment{Backprop \& Step}
			
			\If{IsRefinementIteration($i$)}
			\ForAll{$J$ Gaussians $(p_j, s_j, \alpha_j, c_j, \mathbf{e}_j)$ in $(p, s, a, c, \mathbf{e})$}
			\If{$\alpha < \epsilon$ or IsTooLarge($p_j, s_j)$}	\Comment{Pruning}
			\State RemoveGaussian()	
			\EndIf
			\If{$\nabla_p L > \tau_p$} \Comment{Densification}
			\If{$\|S\| > \tau_S$}	\Comment{Over-reconstruction}
			\State SplitGaussian($p_j, s_j, \alpha_j, c_j, e_j$)
			\Else								\Comment{Under-reconstruction}
			\State CloneGaussian($p_j, s_j, \alpha_j, c_j, e_j$)
			\EndIf	
			\EndIf
			\EndFor		
			\EndIf
			\State $i \gets i+1$
			\EndWhile
		\end{algorithmic}
            \label{alg:gaussian_alg}
 % \end{algorithm*}
 \end{algorithm}

The process of Gaussian Grouping is illustrated in Figure~\ref{fig:model}. We outline the pseudocode for our Gaussian Grouping in Algorithm~\ref{alg:gaussian_alg}, where we highlight the introduced core components in both red and bold texts. 

\vspace{-0.2in}
\subsubsection{(a) 2D Image and Mask Input} To prepare the input for Gaussian Grouping, in Figure~\ref{fig:model}(a), we first deploy SAM to automatically generate masks for each image of the multi-view collection. The 2D masks are individually produced per image. Then, to assign each 2D mask a unique ID in the 3D scene, we need to associate the masks of the same identity across different views and obtain the total number $K$ of instances / stuff in the 3D scene.

\vspace{-0.2in}
\subsubsection{(b) Identity Consistency across Views} Instead of resorting to the cost-based linear assignment~\cite{siddiqui2023panoptic} during training, we treat the multi-view images of a 3D scene as a video sequence with gradually changing views. 
To achieve the 2D mask consistency across views, we employ a well-trained zero-shot tracker~\cite{cheng2023tracking} to propagate and associate masks. 
This also provides the total number of mask identities in the 3D scene.
We visualize associated 2D mask labels in Figure~\ref{fig:model}(b). 
Compared to the cost-based linear assignment proposed in~\cite{siddiqui2023panoptic}, we find it simplifies the training difficulty while avoiding repeatedly computing the matching relation in each rendering iteration, resulting in over 60$\times$ speeding up.
It also achieves better performance than the cost-based linear assignment, especially under the dense and overlapping masks by SAM. Besides, we show the robustness of our 3D masks association in Figure~\ref{fig:correction}, where the 2D associated masks~\cite{cheng2023tracking} from video contain obvious errors. 

\vspace{-0.2in}

% \subsubsection{Identity Encoding} 
\subsubsection{(c) 3D Gaussian Rendering and Grouping} 
To generate 3D-consistent mask identities across views of the scene, we propose to group 3D Gaussians belonging to the same instance / stuff. 
In addition to the existing Gaussian properties, we introduce a new parameter, i.e., Identity Encoding, to each Gaussian. 
The Identity Encoding is a learnable and compact vector of length 16, which we find is sufficient to distinguish the different objects / parts in the scene with computation efficiency.
During training, similar to Spherical Harmonic (SH) coefficients representing the color of each Gaussian, we optimize the introduced Identity Encoding vector to represent its instance ID of the scene.
Note that different from the view-dependent appearance modeling of the scene, the instance ID is consistent across various rendering views.
Thus, we set the SH degree of the Identity Encoding to 0, to only model its direct-current component.
Unlike NeRF-based methods~\cite{siddiqui2023panoptic,kobayashi2022decomposing,lerf2023} designing extra semantic MLP layers, the Identity Encoding serves as the learnable property for each Gaussian to group the 3D scene.

To optimize the introduced Identity Encoding of each Gaussian, in Figure~\ref{fig:model}(c), we render these encoded identity vectors into 2D images in a differentiable manner. 
We take the differentiable 3D Gaussian renderer from~\cite{kerbl20233d} and treat the rendering process similar to the color (SH coefficients) optimization in~\cite{kerbl20233d}.  

3D Gaussian splatting adopts neural point-based $\alpha'$-rendering~\cite{kopanas2021point,kopanas2022neural}, where the influence weight $\alpha'$ of each Gaussian can be evaluated in 2D for each pixel.
Following~\cite{kerbl20233d}, the influence of all Gaussians on a single pixel location is computed by sorting the Gaussians in depth order and blending $\mathcal{N}$ ordered points
overlapping the pixels~\cite{max1995optical,mildenhall2020nerf}:
\begin{equation}
\label{eq:2d_id}
E_{\text{id}} = \sum_{i \in \mathcal{N}} e_i \alpha'_i \prod_{j=1}^{i-1} (1 - \alpha'_j)
\end{equation}
where the final rendered 2D mask identity feature $E_{\text{id}}$ for each pixel is a weighted sum over the Identity Encoding $e_i$ of length 16 for each Gaussian, weighted by the Gaussian's influence factor $\alpha'_i$ on that pixel. Refer to~\cite{yifan2019differentiable}, we compute $\alpha'_i$ by measuring a 2D Gaussian with covariance $\Sigma^{\textrm{2D}}$ multiplied with a learned per-point opacity $\alpha_i$, and
\begin{equation}
\Sigma^{\textrm{2D}} = J W \Sigma^{\textrm{3D}} W^T J^T
\end{equation}
where $\Sigma^{\textrm{3D}}$ is the 3D covariance matrix, $\Sigma^{\textrm{2D}}$ is the splatted 2D version~\cite{zwicker2001surface}. 
$J$ is the Jacobian of the affine approximation of the 3D-2D projection, and $W$ is the world-to-camera transformation matrix. 

\subsubsection{(d) Grouping Loss} 
After associating 2D instance labels across each training view, suppose there are $K$ masks in total in the 3D scene.
To group each 3D Gaussian by the instance /stuff mask identities, we design the grouping loss $\mathcal{L}_\text{id}$ for updating the Identity Encoding of Gaussians with two components:

1.~\textit{\textbf{2D Identity Loss}}:
Since the mask identity labels are in 2D, instead of directly supervising the Identity Encoding $e_i$ of the 3D Gaussians. 
Given the rendered 2D features $E_{\text{id}}$ in Eq.~\ref{eq:2d_id} as input, we first add a linear layer $f$ to recover its feature dimension back to $K$ and then take $\mathit{softmax}(f(E_{\text{id}}))$ for identity classification, where $K$ is the total number of masks in the 3D scene. We adopt a standard cross-entropy loss $\mathcal{L}_\text{2d}$ for $K$ categories classification.

2.~\textit{\textbf{3D Regularization Loss}}: 
To further boost the grouping accuracy of Gaussians, besides the standard cross-entropy loss for indirect 2D supervision, we also introduce an unsupervised 3D Regularization Loss to directly regularize the learning of Identity Encoding $e_i$.
3D Regularization Loss leverages the 3D spatial consistency, which enforces the Identity Encodings of the top $k$-nearest 3D Gaussians to be close in their feature distance. 
This allows the 3D Gaussians inside the 3D object, or heavily occluded (not visible in nearly all training views) during the point-based rendering (Eq.~\ref{eq:2d_id}) to be supervised more sufficiently. 
In Eq.~\ref{eq:reg_3d}, we denote $F$ as softmax operation combined after linear layer $f$ (shared in computing the 2D Identity loss). 
We formalize the KL divergence loss with $m$ sampling points as,
\begin{equation}
\label{eq:reg_3d}
\mathcal{L}_\text{3d} = \frac{1}{m} \sum_{j=1}^{m} D_{\text{kl}}(P || Q) = \frac{1}{mk} \sum_{j=1}^{m}\sum_{i=1}^{k} F(e_j) \log\left(\frac{F(e_j)}{F(e'_i)}\right)
\end{equation}
where $P$ contains the sampled Identity Encoding $e$ of a 3D Gaussian, while the set $Q = \{e'_1, e'_2, ..., e'_k\}$ consists of its $k$ nearest neighbors in 3D Euclidean space. We omit the softmax operation combined after linear layer $f$ for simplicity.

Combined with the conventional 3D Gaussian Reconstruction Loss $\mathcal{L}_\text{rec}$ on image rendering~\cite{kerbl20233d}, the total loss $\mathcal{L}_\text{render}$ for fully end-to-end training is
\begin{equation}
    \mathcal{L}_\text{render}  = \mathcal{L}_\text{rec} + \mathcal{L}_\text{id} = \mathcal{L}_\text{rec} + \lambda_\text{2d}\mathcal{L}_\text{2d} + \lambda_\text{3d}\mathcal{L}_\text{3d}
\end{equation}

\vspace{-0.1in}
\begin{figure}[!t]
	\centering
	\vspace{-0.2in}
	\includegraphics[width=1.0\linewidth]{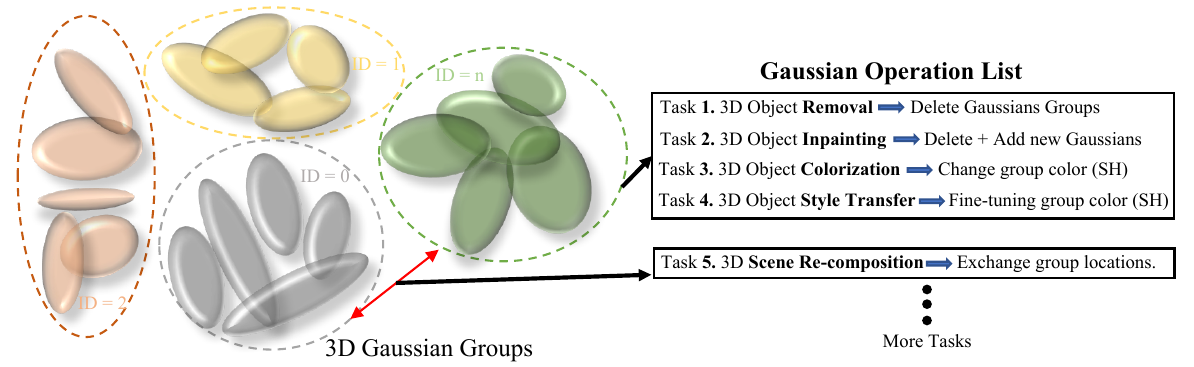}
	\vspace{-0.25in}
	\caption{The grouped 3D Gaussians after training, where each group represents a specific instance / stuff of the 3D scene and can be fully decoupled. Our representation is efficient to support versatile downstream scene editing applications, where we design a Gaussian Operation List consisting of simple operations like group deletion, group addition, finetuning Spherical Harmonic (SH) and exchanging 3D center locations.}
	\label{fig:gaussian_group}
	\vspace{-0.2in}
\end{figure}

\vspace{-0.1in}
\subsection{Gaussian Grouping for Scene Editing}
\label{sec:partial_tuning}

After the 3D Gaussian field training and grouping (Sec~\ref{sec:gau_group}), as in Figure~\ref{fig:gaussian_group}, we represent the whole 3D scene with a set of grouped 3D Gaussians.
To perform various downstream local scene editing tasks, we propose efficient Local Gaussian Editing.
Thanks to the decoupled scene representation, instead of fine-tuning all 3D Gaussians, we freeze the properties for most of the well-trained Gaussians and only adjust a small part of existing or newly added 3D Gaussians relevant to the editing targets.
For 3D object removal, we simply delete the 3D Gaussians of the editing target.
For 3D scene re-composition, we exchange the 3D location between two Gaussian groups.
These two editing applications are direct with no parameter tuning. For 3D object inpainting, we first delete the relevant 3D Gaussians and then add a small number of new Gaussians to be supervised by the 2D inpainting results by LaMa~\cite{lama} during rendering.
For 3D object colorization, we only tune the color (SH) parameters of the corresponding Gaussian group to preserve the learned 3D scene geometry. For 3D object style transfer, we further unfreeze the 3D positions and sizes to achieve more realistic results.

Our local Gaussian editing scheme is time-efficient, as shown in our experiment. Because of our fine-grained mask modeling, it also supports multiple concurrent local editings without interfering with each other or re-training the whole global 3D scene representation with new editing operations. Compared to NeRF-based approaches~\cite{mirzaei2023spin,kobayashi2022decomposing,instructnerf2023}, we also provide extensive visual comparisons for scene editing cases and discussion on mask granularity in the Supp. file.
\vspace{-0.2in}

\section{Experiments}
\label{sec:exp}

\subsection{Dataset and Experiment Setup}

\subsubsection{Datasets} 
To measure segmentation or fine-grained localization accuracy in open-world scene, we evolve the existing LERF-Localization~\cite{lerf2023} evaluation dataset and propose the \textit{LERF-Mask} dataset, where we manually annotate three scenes from LERF-Localization with accurate masks instead of using coarse bounding boxes. For each 3D scene, we provide 7.7 text queries with corresponding GT mask labels on average. We also provide 3D panoptic segmentation results on Replica~\cite{replica19arxiv} and ScanNet~\cite{dai2017scannet} dataset.
More details are in the Supp. file

To evaluate the reconstruction quality, we tested our Gaussian Grouping on 7 of full 9 sets of scenes presented in Mip-NeRF 360~\cite{barron2022mipnerf360}, where the flowers and treehill are skipped due to the non-public access right. We also take diverse 3D scene cases from LLFF~\cite{mildenhall2019llff}, Tanks \& Temples~\cite{knapitsch2017tanks} and Instruct-NeRF2NeRF~\cite{instructnerf2023} for visual comparison on scene editing.
\vspace{-0.1in}

\subsubsection{Implementation Details} 
We implement Gaussian Grouping based on Gaussian Splatting~\cite{kerbl20233d}. We add a 16-dimension identity encoding to each Gaussian and implement forward and backward cuda rasterization similar to the RGB feature. The output classification linear layer has 16 input and 256 output channels. In training, $\lambda_\text{2d} = 1.0$ and $\lambda_\text{2d} = 2.0$.  We use the Adam optimizer for both Gaussians and the linear layer, with a learning rate of 2.5e-3 for identity encoding and 5e-4 for the linear layer. For 3D regularization loss, we choose $k=5$ and $m=1000$. All datasets are trained for 30K iterations on one A100 GPU. Refer to our Supp. file for more details on scene editing and SAM's mask granularity.

\begin{figure}[!t]
    \begin{minipage}[t]{0.46\linewidth}
	\centering
	\vspace{-0.1in}
	\includegraphics[width=1.0\linewidth]{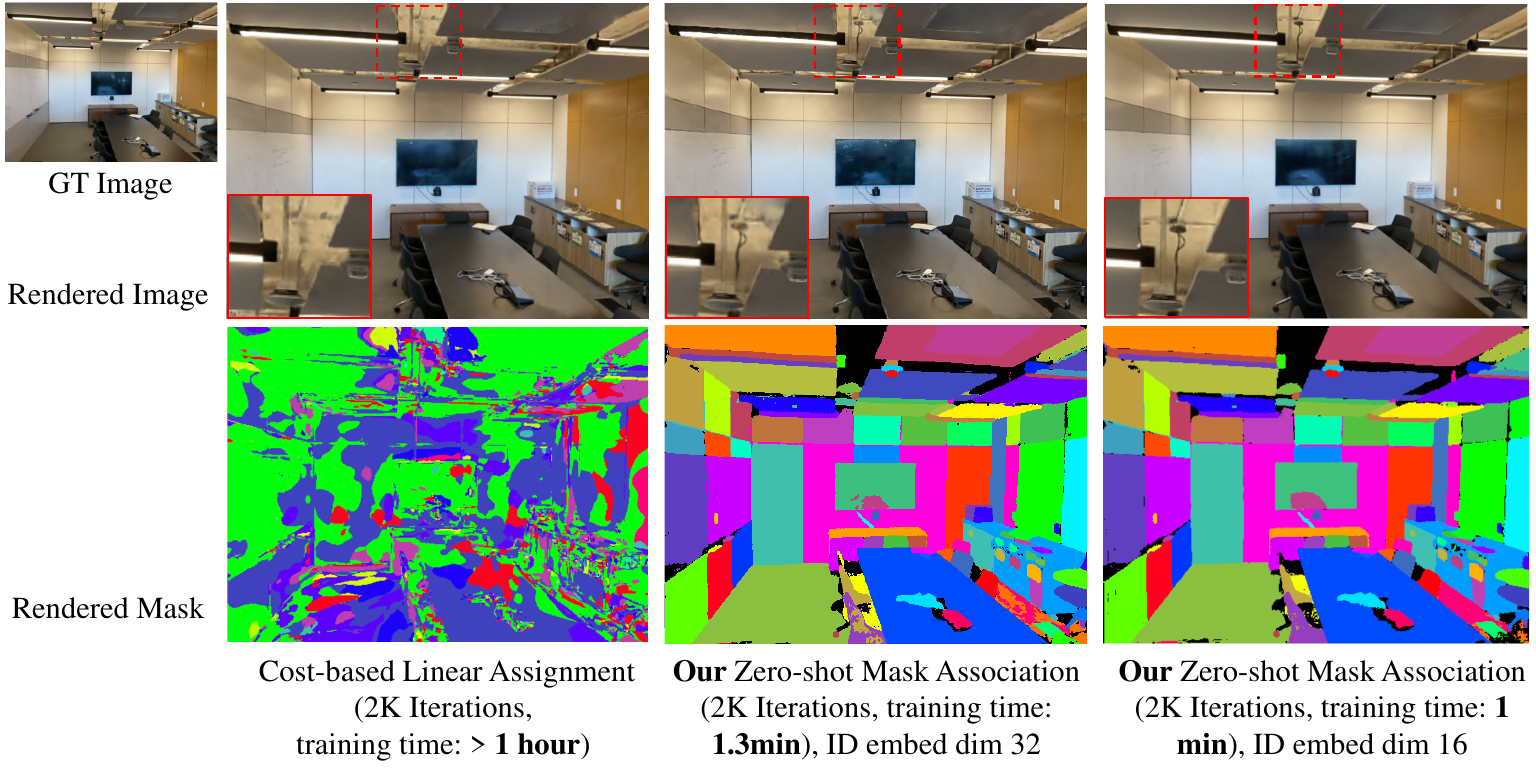}
	\vspace{-0.2in}
	\caption{Ablation on the Identity Consistency across views, where we treat multi-view images as a video and associate the mask labels to generate coherent segmentation labels~\cite{cheng2023tracking} for training. We founding using cost-based linear assignment~\cite{siddiqui2023panoptic} leads to slower training and inferior testing performance in both reconstruction and segmentation.} 
	\label{fig:deva_ablation}
	\vspace{-0.2in}
    \end{minipage}
    \hfill
    \begin{minipage}[t]{0.52\linewidth}
    
    \centering
	\vspace{-0.1in}
	\includegraphics[width=1.0\linewidth]{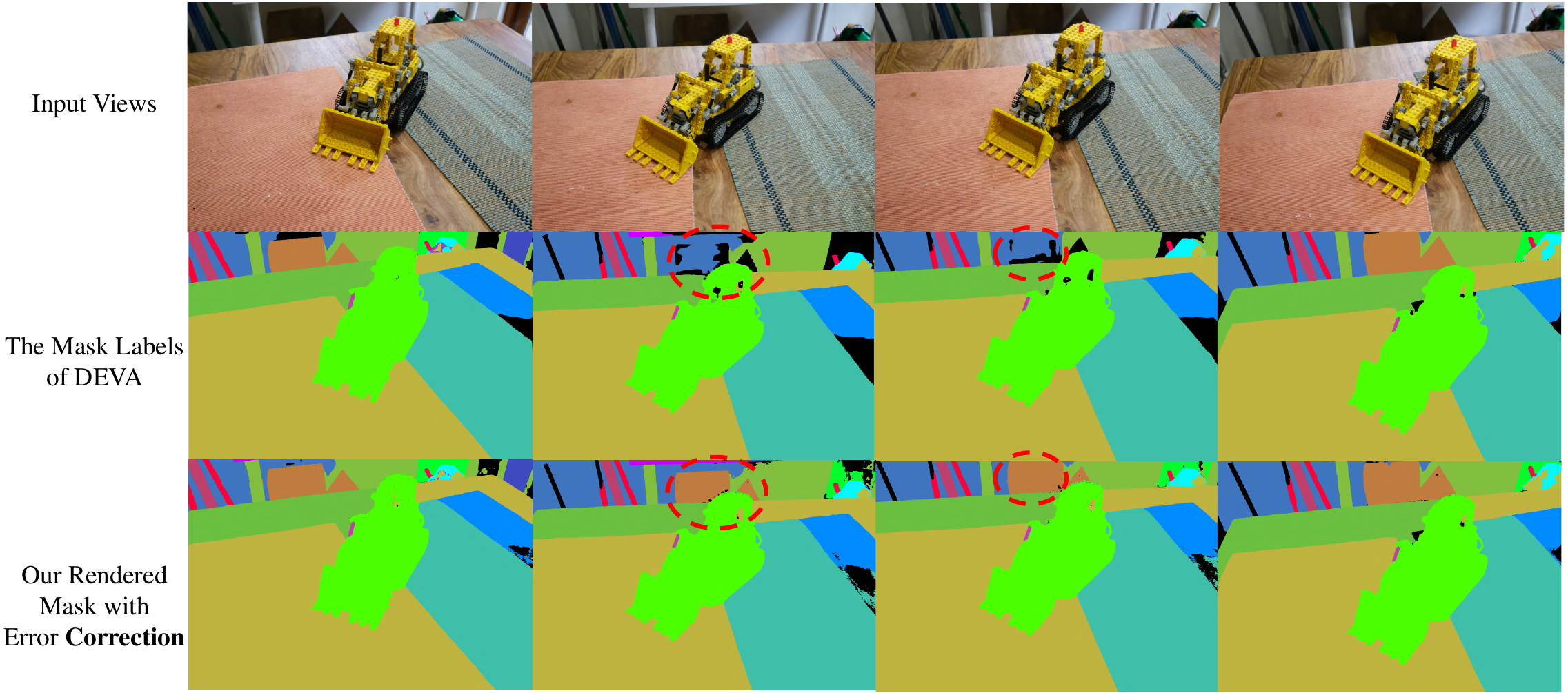}
	\vspace{-0.2in}
	\caption{Robustness to input masks errors on Mip-NeRF 360~\cite{barron2022mipnerf360}. In the 2nd and 3rd columns (middle two views), SAM + DEVA fails to segment and associate the chair across frames. However, owing to the shared 3D Gaussian representation during reconstruction, Gaussian Grouping successfully corrects the error in mask labels and segments the black chair during the multi-view rendering.}%
	\label{fig:correction}%
	\vspace{-0.2in}
    \end{minipage}
\end{figure}

\subsection{Ablation Experiments}

\subsubsection{Ablation on Mask Cross-view Association} To study the effect of cross-view masks association~\cite{cheng2023tracking} for input preparation, we replace the associated masks input to the individual masks predicted by SAM per image in Gaussian Grouping. We take the cost-based linear assignment strategy proposed in~\cite{siddiqui2023panoptic}, and perform the visual comparison on rendering results in Figure~\ref{fig:deva_ablation}. The linear assignment not only heavily slows down the whole training due to the assignment computation in each iteration, but also produces noisy mask predictions. 
This is owing to the large gradients brought by the unstable cost-based liner assignment, especially at the initial training stage of the network, where masks prediction is nearly random.
For 2K training iteration, linear assignment requires 1 hour but our associated mask input only requires 1 minute.
Also, we compare the scene render reconstruction quality in Figure~\ref{fig:deva_ablation}, where the appearance details of the rope attached to the ceiling are much better preserved by our method.

\vspace{-0.1in}
\subsubsection{Masks Association Errors Correction and Robustness}
Gaussian Grouping can also correct the 2D segmentation errors produced by DEVA, as shown in Figure~\ref{fig:correction}. Even taking the training input with partial 2D mask labels lost by tracking, Gaussian Grouping is robust in 3D association and can retrieve them based on the shared 3D Gaussian representation across different views.

\begin{table}[!t]
			\centering%
		\begin{minipage}[t]{0.62\linewidth}
			\centering
\caption{Influence of Identity Encoding on Mip-NeRF 360~\cite{barron2022mipnerf360} dataset with its 7 public scenes. The joint training of the introduced Identity Encodings does not hurt the original Gaussian reconstruction quality.} 
\vspace{-0.1in}
\resizebox{0.92\linewidth}{!}{%
\begin{tabular}{l|c|c|ccc|c}
\hline
Model & Scene Seg & Scene Edit & \multicolumn{1}{c}{PSNR$\uparrow$} & \multicolumn{1}{c}{SSIM$\uparrow$} & \multicolumn{1}{c|}{LPIPS$\downarrow$} & \multicolumn{1}{c}{FPS} \\ \hline
% M-NeRF360~\cite{barron2022mipnerf360} &  & 29.09 & & & \\
Baseline: Gaussian Splatting~\cite{kerbl20233d} & - & - & 28.69 & 0.870 & 0.182 & $\sim$200 \\
\textbf{Gaussian Grouping} & \textbf{$\checkmark$} & \textbf{$\checkmark$} & 28.43 & 0.863 & 0.189 & $\sim$170 \\ \hline
\end{tabular}
}\label{tab:identity_ablation}\vspace{-0.1in}
		\end{minipage}
		\label{tab:test1}
		\hfill
		\begin{minipage}[t]{0.35\linewidth}
			\centering
\small
\caption{Ablation of K of 3D Regularization Loss on the 3D object removal. RAcc: Object Removal Accuracy.}
\vspace{-0.13in}
\resizebox{1.0\linewidth}{!}{
\begin{tabular}{l|c|ccccc}
\toprule
   \multirow{2}{*}{Model} & \multirow{2}{*}{Gaussian Splatting} & \multicolumn{5}{|c}{Gaussian Grouping}  \\ 
  & & K=0 & K=1 & k=2 & K=5 & K=10     \\ \midrule  
PSNR & 30.32 & 30.51 & 30.62 & 30.61 & \textbf{30.72} & 30.62  \\ 
RAcc & N/A & 41.2\% & 40.5\% & 67.5\% &76.6\% & \textbf{77.8}\% \\
 \bottomrule
\end{tabular}}
% \vspace{-0.25in}
\label{exp:k}
		\end{minipage}\vspace{-0.1in}%
		%\hspace{\fill}
	\end{table}

\vspace{-0.1in}

\subsubsection{Influence of the Identity Encoding}
In Table~\ref{tab:identity_ablation}, we study the influence of our introduced Identity Encoding on original Gaussian Splatting's 3D reconstruction performance and speed. The performance of Gaussian Grouping is on par with the original Gaussian Splatting method with negligible decrease, but it allows anything in the whole 3D scene to be segmented. By Identity Encoding, we lift the 2D zero-shot segmentation of SAM to the 3D open world. 
Further, the grouped 3D Gaussians support a wide range of downstream editing tasks. We simply train all model components in an end-to-end manner jointly. This is different from Panoptic Lifting~\cite{kundu2022panoptic} which requires to block gradients from the segmentation branch back to the reconstruction branch.

\vspace{-0.1in}
\subsubsection{Ablation on Identity Encoding Dimension}
We study the impact of dimension for our Identity Encoding in Figure~\ref{fig:deva_ablation}.
To keep the training efficiency, we set the dimension of Identity Encoding to only 16 because it not only shows a good segmentation separation between objects but also keeps the training efficiency.
Doubling the dimension to 32 does not bring a better reconstruction quality compared to 16 but make training 1.3 times slower.

\begin{figure}[!t]
\centering%
% \vspace{0.05in}
\includegraphics[width=1.0\linewidth]{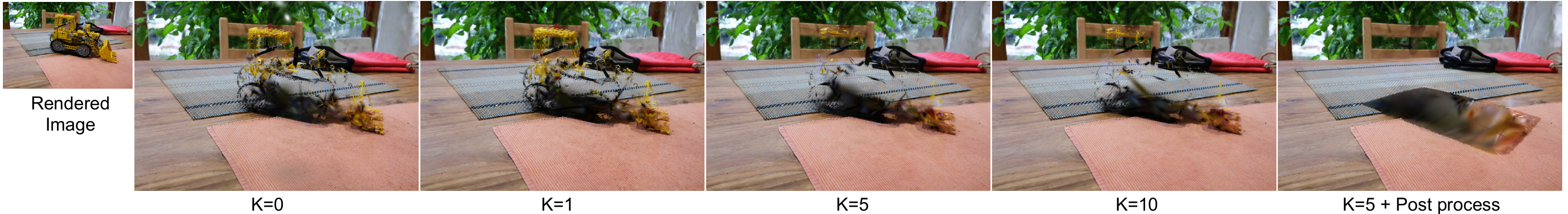}\vspace{-0.12in}%
\caption{Visual ablation of K in the 3D Regularization Loss on object removal editing of MipNeRF360. We remove Gaussians classified as lego with various K. We compute the convex hull of the removed 3D Gaussian points as the post process.}\vspace{-0.15in}%
\label{fig:ablation_k}%
% \vspace{-0.05in}
\end{figure}

\begin{figure}[!t]
	\centering
	% \vspace{-0.1in}
	\includegraphics[width=1.0\linewidth]{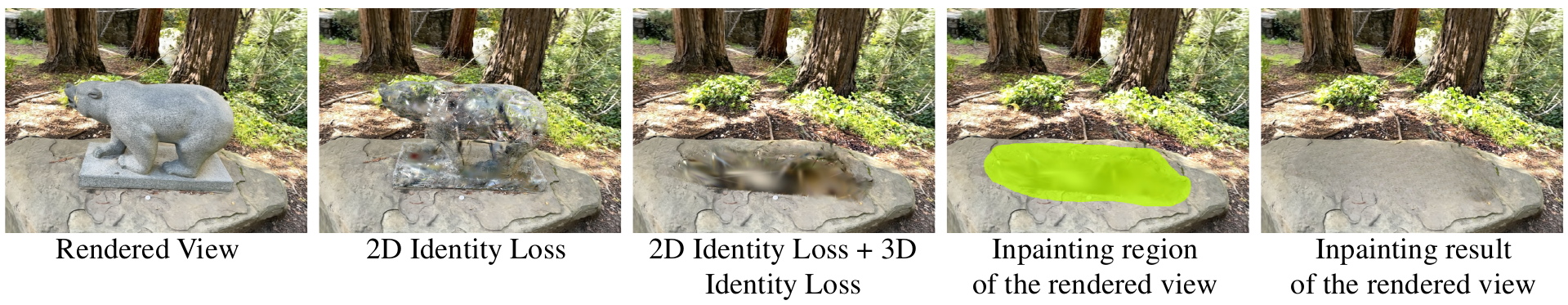}
	\vspace{-0.3in}
	\caption{Ablation on the Grouping Loss on the Bear inpainting case. The joint supervision of 2D and 3D losses addresses the ``transparent bear issue", which shows better Gaussian Grouping accuracy. After deleting the Gaussians belonging to the bear, we detect the image hole region with no Gaussians projection covering and then use LaMa~\cite{lama} to produce a single view inpainting result to guide the learning for newly added inpainting Gaussians.}
	\label{fig:inpaint_ablation}
	\vspace{-0.25in}
\end{figure}

\vspace{-0.2in}
\subsubsection{Ablation of 3D Regularization Loss} We perform ablation of K in our 3D Regularization Loss on the Kitchen dataset of Mip-NeRF 360~\cite{barron2022mipnerf360} to select K. As in Fig.~\ref{fig:ablation_k} and Table~\ref{exp:k}, $K=5$ achieves both the best balance between the scene reconstruction and 3D object removal accuracy. 
Figure~\ref{fig:ablation_k} further visualizes the influence of K during the 3D object removal process.

\vspace{-0.2in}

\subsubsection{Visual Ablation on the Grouping Losses}
In Figure~\ref{fig:inpaint_ablation}, we study the effect of our grouping loss components, where solely using 2D Identity Loss will result in the `transparent bear issue’. This is due to Gaussians inside the bear being occluded during training and cannot be supervised sufficiently. We address it by proposing 3D Identity/Regularization Loss for joint training with the 2D loss.
\vspace{-0.1in}

\subsection{3D Multi-view Segmentation}

\subsubsection{Open-vocabulary Segmentation Comparison} 
\begin{figure*}[!t]
	\centering
	\vspace{-0.05in}
	\includegraphics[width=1.0\linewidth]{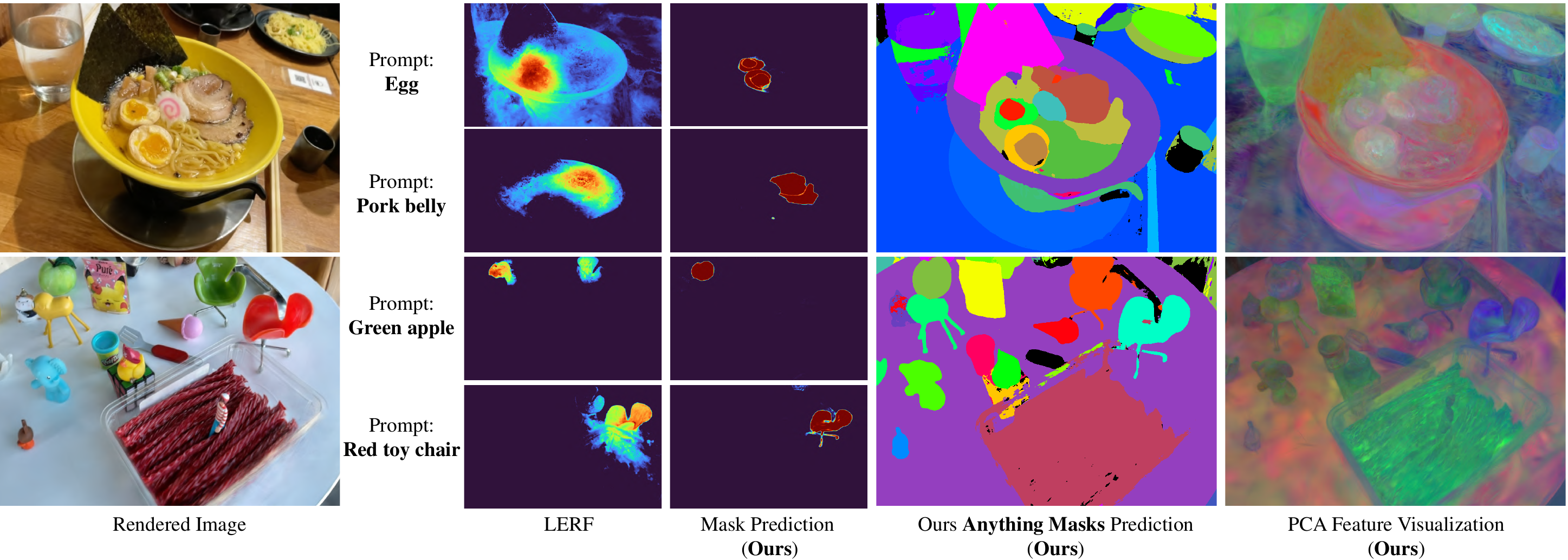}
	\vspace{-0.28in}
	\caption{Segmentation comparison between LERF~\cite{lerf2023} and our Gaussian Grouping on the rendering view. The masks predicted by Gaussian Grouping contain much sharper and more accurate boundaries than LERF. Also, our approach is better at distinguishing objects with similar colors, such as the ``Green apple" prompt case.}
	\label{fig:seg_c}
	\vspace{-0.25in}
\end{figure*}

We compare the segmentation quality of Gaussian Grouping in 3D scenes with the state-of-the-art open-vocabulary 3D segmentation methods, such as LERF~\cite{lerf2023} and SA3D~\cite{cen2023segment}. 
To compare fine-grained mask localization quality, we annotate the test views of three 3D scenes from the LERF-Localization~\cite{lerf2023} dataset with accurate masks to replace its original coarse bounding boxes.
In Table~\ref{tab:seg_comp}, the advantage of our Gaussian Grouping is significant, doubling the performance of LERF and SA3D on both the "figurines" and "ramen" scenes. We also show the visual segmentation comparison in Figure~\ref{fig:seg_c}, where the segmentation prediction by our method is much more accurate with a clear boundary, while LERF-based similarity computing only provides a rough localization area. Since SAM does not support language prompts, both SA3D and our method adopt Grounding DINO~\cite{liu2023grounding} to identify the mask ID in a 2D image, and then pick the corresponding mask from our anything masks prediction. 

\vspace{-0.1in}

\subsubsection{3D Panoptic Segmentation Comparison} We compare the panoptic segmentation quality with Panoptic Lifting \cite{siddiqui2023panoptic} in Table~\ref{tab:panoptic}. We use the same semantic mask labels from Mask-DINO~\cite{li2022mask} with semantic information for both Panoptic Lifting~\cite{siddiqui2023panoptic} and our method. Gaussian Grouping outperforms Panoptic Lifting in both performance and speed. 

\vspace{-0.1in}
\subsection{3D Scene Editing}
\subsubsection{3D Object Removal}
\begin{figure}[!t]
	\centering
	\vspace{-0.1in}
	\includegraphics[width=1.0\linewidth]{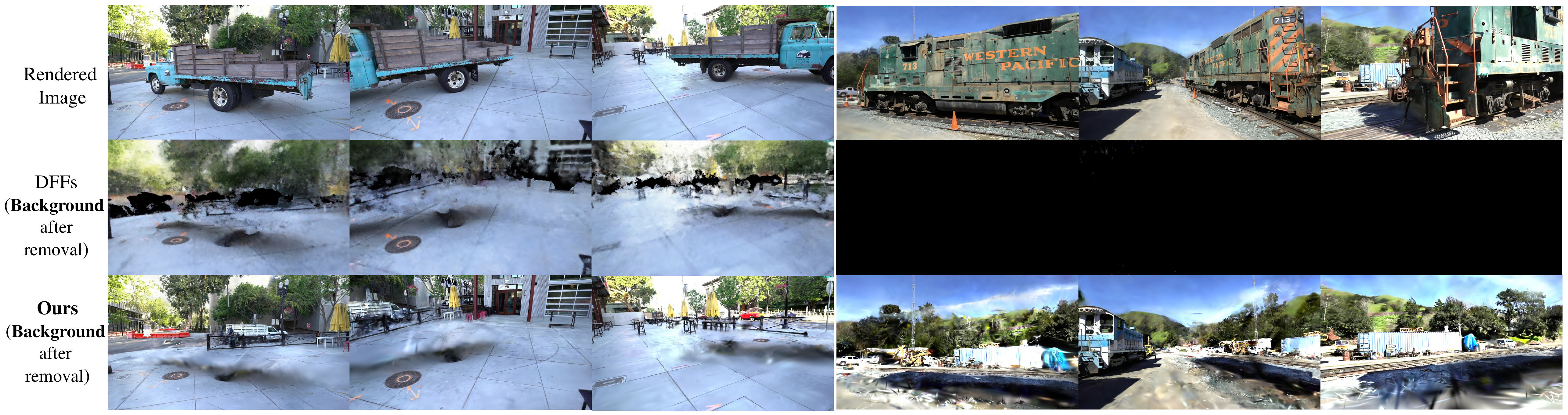}
	\vspace{-0.2in}
	\caption{3D Object removal on the Tanks \& Temples dataset~\cite{knapitsch2017tanks}. Compared to DFFs~\cite{kobayashi2022decomposing}, our Gaussian Grouping can remove the large-scale objects, such as truck, from the 3D scene with greatly reduced artifacts \textbf{w/o} leaving a blurry background.}%
	\label{fig:removal}%
	\vspace{-0.03in}
\end{figure}

3D object removal is to completely delete an object from the 3D scene, where the background behind the removed instance / stuff can be noisy or have a hole because of no observation. In Figure~\ref{fig:removal}, we compare the removal effect of our Gaussian Grouping with the Distilled Feature Fields (DFFs)~\cite{kobayashi2022decomposing}. For challenging scene cases with large objects, our method can clearly separate the 3D object from the background. While the performance of DFFs is limited by the quality of its CLIP-distilled features, which results in the complete foreground removal (Train case) or inaccurate region removal with obvious artifacts (Truck case).
\vspace{-0.15in}

\begin{table}[!t]
\vspace{0.1in}
			\centering%
		\begin{minipage}[t]{0.52\linewidth}
			\centering
\small
\caption{Comparison of Open Vocabulary Segmentation on LERF-Mask dataset.
We adopt the detections from Grounding DINO~\cite{liu2023grounding} to select mask IDs in a 3D scene.}
\vspace{-0.13in}
\resizebox{0.95\linewidth}{!}{
\begin{tabular}{l|cc|cc|cc}
\toprule
   \multirow{2}{*}{Model} & \multicolumn{2}{c|}{figurines} & \multicolumn{2}{c|}{ramen} & \multicolumn{2}{c}{teatime}  \\ 
  & mIoU & mBIoU & mIoU & mBIoU & mIoU & mBIoU    \\ \midrule 
DEVA~\cite{cheng2023tracking}&46.2 & 45.1 & 56.8  & 51.1 & 54.3 & 52.2 \\
LERF~\cite{lerf2023} & 33.5 & 30.6 & 28.3 & 14.7 & 49.7 & 42.6  \\   
SA3D~\cite{cen2023segment} & 24.9 & 23.8 & 7.4 & 7.0 & 42.5 & 39.2 \\ 
LangSplat~\cite{qin2023langsplat} & 52.8 & 50.5 & 50.4 & 44.7 & 69.5 & 65.6 \\
Gaussian Grouping & \textbf{69.7} & \textbf{67.9} & \textbf{77.0} & \textbf{68.7} & \textbf{71.7} & \textbf{66.1} \\ 
 \bottomrule
\end{tabular}}
% \vspace{-0.05in}
\label{tab:seg_comp}
		\end{minipage}
		\label{tab:test1}
		\hfill
		\begin{minipage}[t]{0.46\linewidth}
			\centering
\small
\caption{Panoptic segmentation comparison on novel views. For semantic information of the prediction, we adopt Mask-DINO~\cite{li2022mask} as the semantic mask label generator for each view. We refer to the reported setting from~\cite{dou2024cosseggaussians}.}
\vspace{-0.1in}
\resizebox{1.0\linewidth}{!}{
\begin{tabular}{l|ccc|ccc}
\toprule
   \multirow{2}{*}{Model} & \multicolumn{3}{|c}{Replica} & \multicolumn{3}{|c}{ScanNet}  \\ 
  & mIoU ($\%$) & PQ$^{\text{scene}}$($\%$) & FPS & mIoU($\%$) & PQ$^{\text{scene}}$($\%$) & FPS     \\ \midrule  
Panoptic Lifting~\cite{siddiqui2023panoptic} & 66.22 & 64.34 & $\sim$10 & 67.01 & 60.74 & $\sim$10 \\  
Gaussian Grouping & \textbf{71.15} & \textbf{66.52} & \textbf{$\sim$140} & \textbf{68.70} & \textbf{61.83} & \textbf{$\sim$150} \\
 \bottomrule
\end{tabular}}
% \vspace{-0.1in}
\label{tab:panoptic}
		\end{minipage}
		%\hspace{\fill}
	\end{table}

\begin{figure}[!t]
	\centering
	\vspace{-0.1in}
	\includegraphics[width=1.0\linewidth]{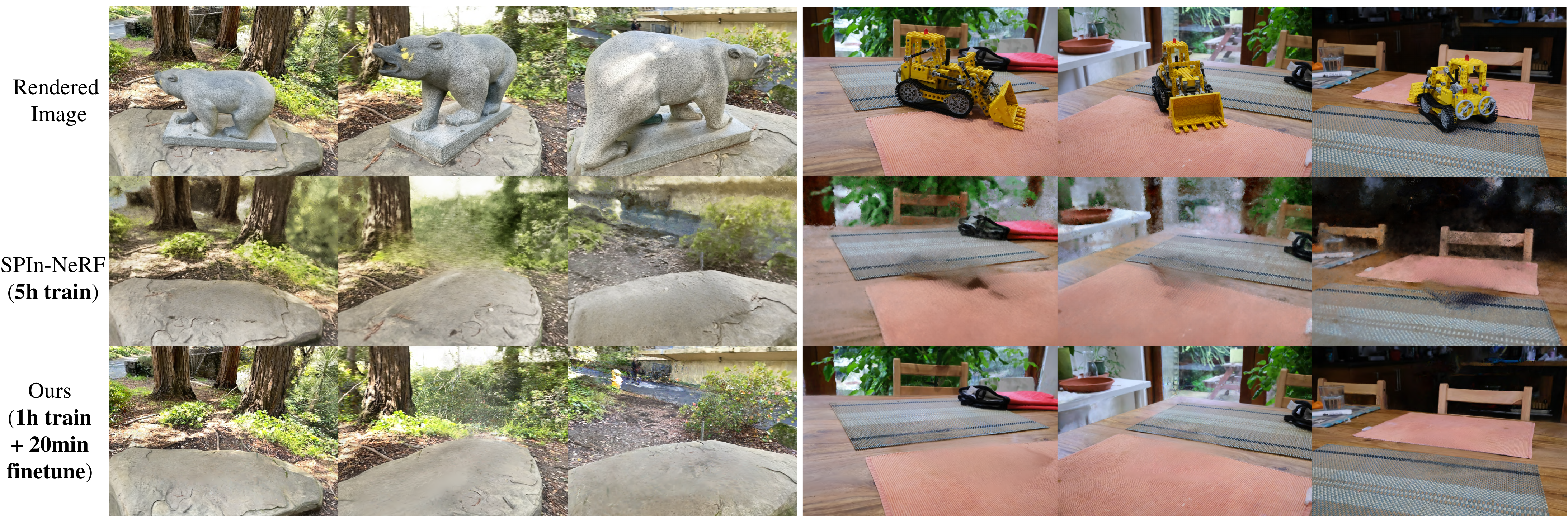}
	\vspace{-0.25in}
	\caption{Comparison on 3D object inpainting cases, where SPIn-NeRF~\cite{mirzaei2023spin} requires 5h training while our method with better inpainting quality only needs 1 hour training and 20 minutes finetuning.}
	\label{fig:inpainting}
	\vspace{-0.12in}
\end{figure}

\begin{figure}[t]
    \vspace{-0.1in}
    \begin{minipage}[t]{0.48\linewidth}
	\centering
	% \vspace{-0.2in}
	\includegraphics[width=1.0\linewidth]{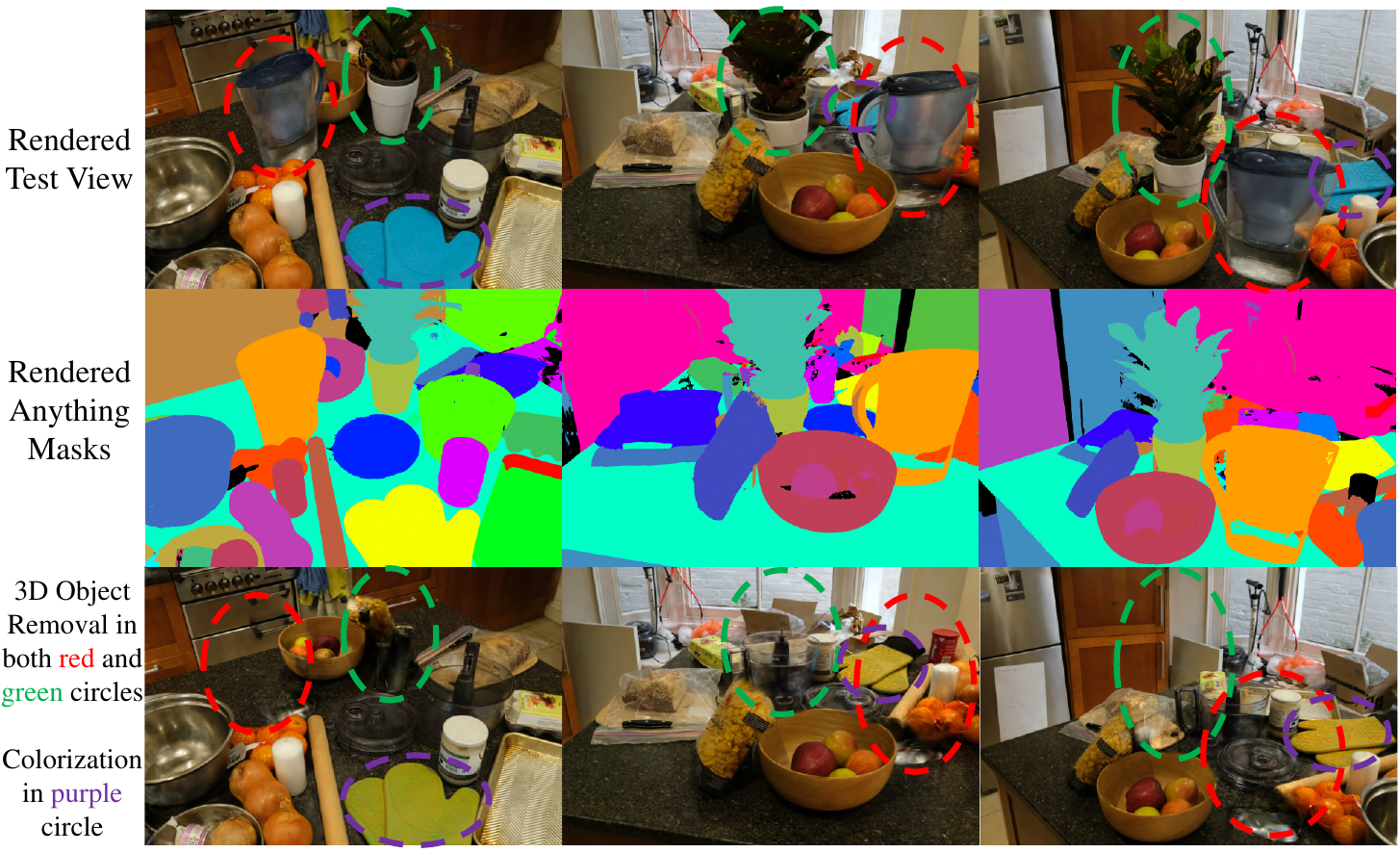}
	\vspace{-0.25in}
	\caption{Multi-object editing within the same 3D scene, where we concurrently perform 3D object removal for objects in the red and green circles, and colorization for the glove in the purple circle.}
	\label{fig:multi_editing}
	% \vspace{-0.25in}
    \end{minipage}
    \hfill
    \begin{minipage}[t]{0.48\linewidth}
    \centering
	% \vspace{-0.1in}
	\includegraphics[width=0.95\linewidth]{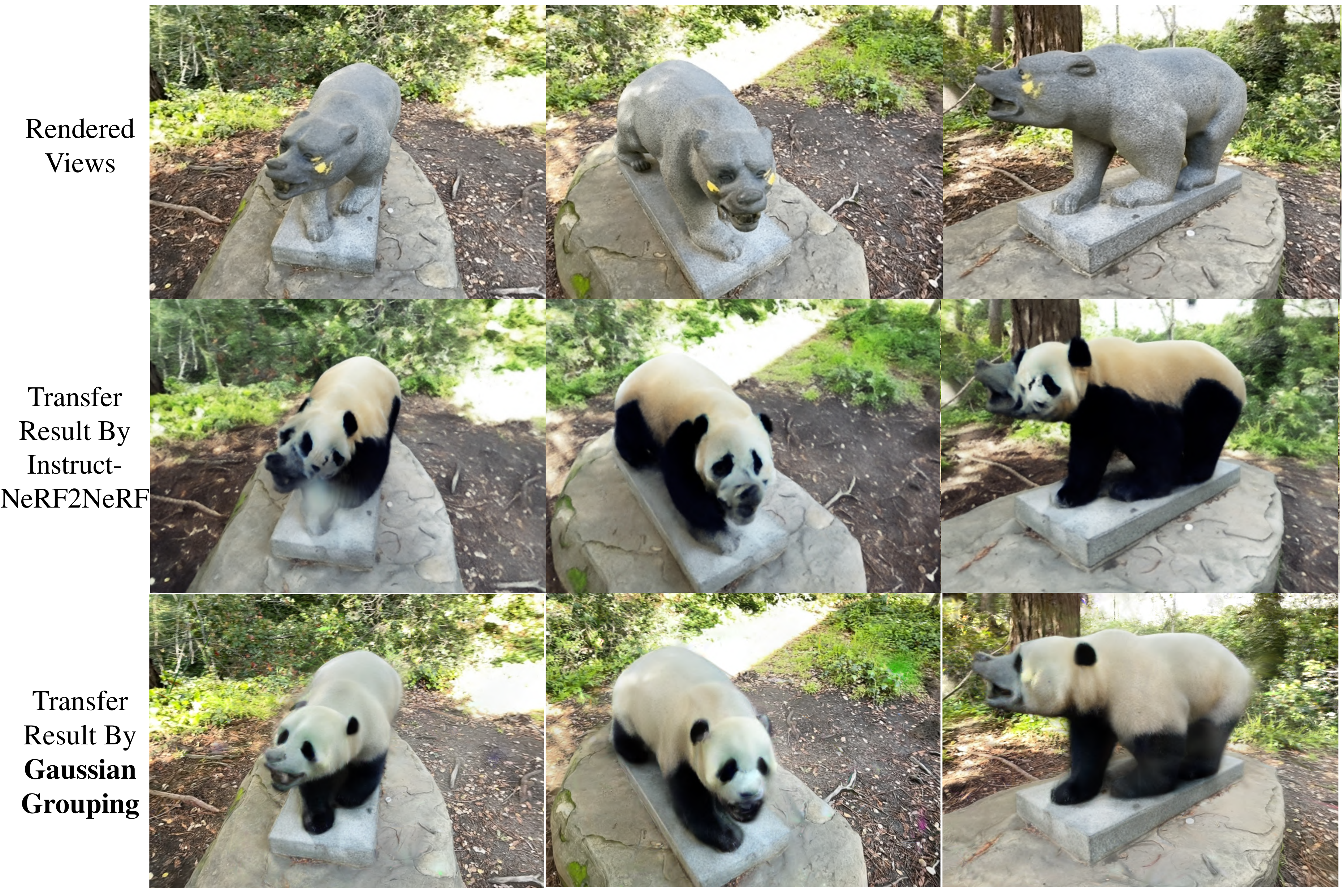}
	\vspace{-0.1in}
	\caption{Visual comparison of 3D object style transfer between Instruct- NeRF2NeRF~\cite{instructnerf2023} and Gaussian Grouping under the same instruction prompt ``Turn the bear into a panda''.} 
	%\label{fig:model}
	\label{fig:style}
	% \vspace{-0.01in}
    \end{minipage}
    \vspace{-0.3in}
\end{figure}

\subsubsection{3D Object Inpainting}
Based on the 3D object removal, in the 3D object inpainting task, we aim to further fill the ``hole regions" due to missing observation across all views and make it a photo-realistic and view-consistent natural 3D scene.
We first detect the regions which are invisible in all views after deletion, and then inpaint these ``invisible regions'' instead of the whole ``2D object regions". 
In each rendering view, we adopt the 2D inpainted image to guide the learning of the newly introduced 3D Gaussian.
In Figure~\ref{fig:inpainting}, compared to SPIn-NeRF~\cite{mirzaei2023spin}, the inpainting result of our Gaussian Grouping better preserves spatial detail
and multi-view coherence.
\vspace{-0.1in}

\subsubsection{3D Object Style Transfer} 

Gaussian Grouping supports 3D object style transfer efficiently.
We compare with the recent Instruct-NeRF2NeRF~\cite{instructnerf2023} on 3D object style transfer in Figure~\ref{fig:style}, using the same instruction prompt ``turn the bear into a panda" and the same image guidance by InstructPix2Pix~\cite{brooks2022instructpix2pix}.
Gaussian Grouping produces more coherent and natural transferred ``bear'' across views.
Since our Gaussian Grouping models ``anything masks'' of the open-world 3D scene, including the bear, the spatial details of background regions (outside the bear) are also faithfully
preserved by our method. While for Instruct-NeRF2NeRF, a large portion of background regions are unnecessarily getting blurry with degraded quality. We describe the detailed pipeline of 3D object style transfer in the Supp. file.

\vspace{-0.1in}

\subsubsection{3D Multi-Object Editing} 
In Figure~\ref{fig:multi_editing}, we demonstrate multiple editing actions (like removing objects and colorization) on different Gaussian groups of the 3D scene. This grouped 3D Gaussian representation enables concurrent editing of several objects while maintaining non-interference among them.

\vspace{-0.1in}

\subsubsection{Limitation} Due to the lack of dynamic modeling and time-dependent updating, the proposed 3D Gaussian Grouping method is currently limited to the static 3D scene. Also, it would also be interesting to further explore fully unsupervised 3D Gaussian grouping in the future. % by removing the 2D mask supervision.
\vspace{-0.1in}

\section{Conclusion}
We propose Gaussian Grouping, the first 3D Gaussian-based approach to jointly reconstruct and segment anything in the open-world 3D scene. 
We introduce an Identity Encoding for 3D Gaussians that is supervised both by 2D mask predictions from SAM and 3D spatial consistency.
Based on this grouped and discrete 3D scene representation, we further show it can support versatile scene editing applications, such as 3D object removal, 3D object inpainting, 3D object style transfer and scene recomposition, with both high-quality visual effects and good time efficiency.

\section{Appendix}

In this supplementary material, we first conduct additional experiment analysis of our Gaussian Grouping in~Section~\ref{sec:supp_exp}, including multi-granularity, segmentation efficiency, quantitative editing evaluation and robustness. 
Then, in Section~\ref{sec:dataset}, we describe the detailed process of annotating our proposed LERF-Mask datasets with the visualization of annotation examples.
We further provide a more detailed 3D object inpainting and style transfer pipeline description in Section~\ref{sec:inpaiting_pipe}.
Finally, we illustrate the algorithm pseudocode of our Gaussian Grouping and more implementation details in~Section~\ref{sec:supp_details}, including the method limitation analysis.
Please refer to our \href{https://ymq2017.github.io/gaussian-grouping}{\textcolor{red}{project page}} for extensive 3D results comparison.

\subsection{Supplementary Experiments}
\label{sec:supp_exp}

\begin{figure*}[!h]
	\centering
	% \vspace{-0.05in}
	\includegraphics[width=1.0\linewidth]{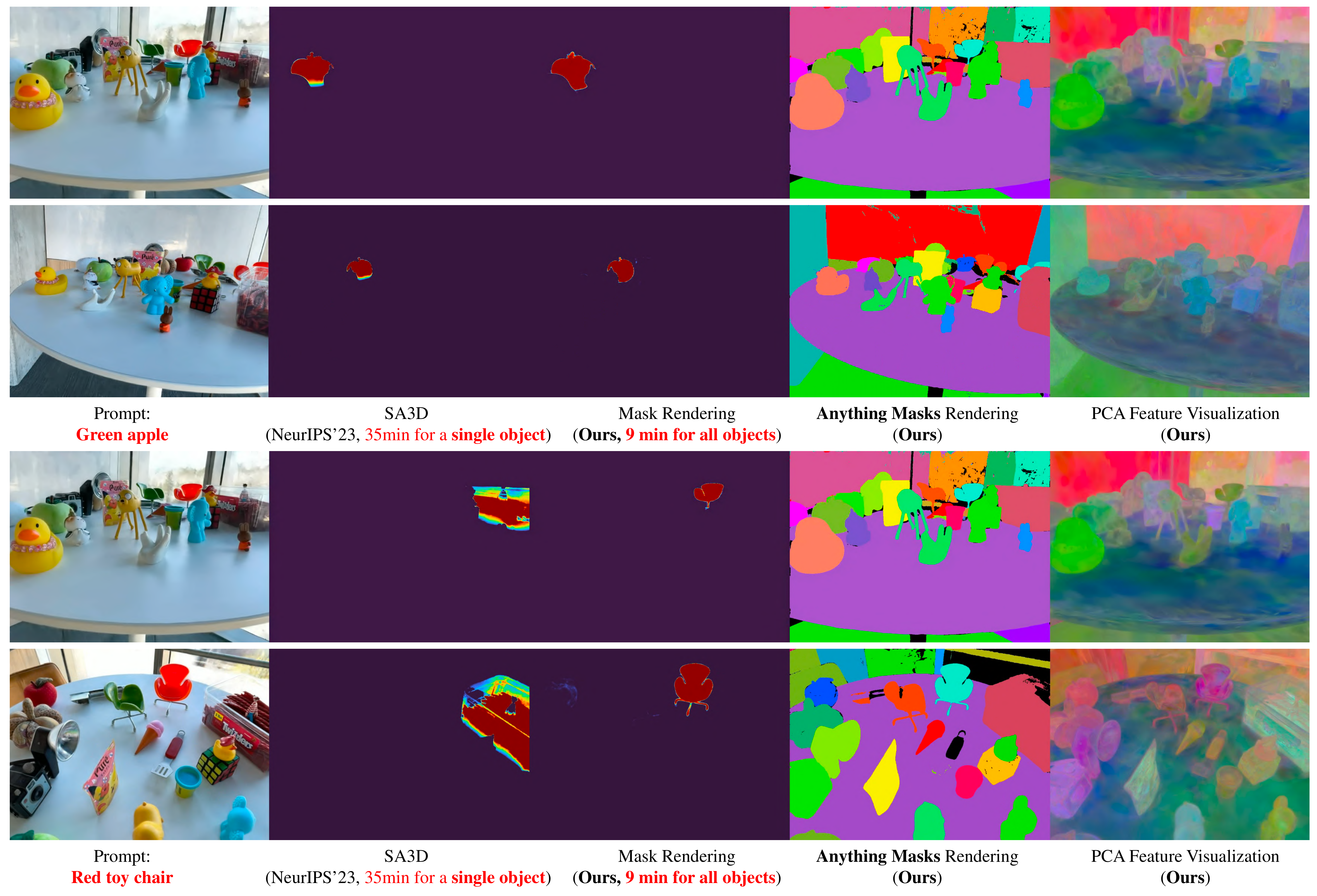}
	\vspace{-0.25in}
	\caption{Segmentation comparison between SA3D~\cite{cen2023segment} and our Gaussian Grouping on the rendering view. We adopt PCA to visualize the rendered Identity Encoding features in the rightmost column. \textit{Note that SA3D does not support concurrent multi-object segmentation} due to its design limitation in the inverse rendering, which requires training and rendering for each segmentation target for around 35 minutes ($\sim$20min for training and $\sim$15min for rendering). In contrast, our Gaussian Grouping shows great efficiency by segmenting all objects in the scene only in 9 minutes.}
	%\label{fig:model}
	\label{fig:sa3d}
	\vspace{-0.01in}
\end{figure*}

\parsection{Segmentation Efficiency}
In Fig.~\ref{fig:sa3d}, we compare the segmentation results with SA3D~\cite{cen2023segment} on the proposed LERF-Mask dataset for open-world 3D segmentation.
Our Gaussian Grouping shows great advantages in segmentation efficiency. Using the same running environment, \textit{our Gaussian Grouping jointly segments \textbf{all objects} of the 3D scene in 9 minutes, while SA3D requires 35 minutes for \textbf{each object}} due to its inverse rendering design in 3D voxel grids.
To segment each object of the 3D scene, SA3D repeatedly needs separate new training, which makes SA3D time-consuming and not user-friendly in multi-object segmentation or editing scenarios.

\begin{figure}[!h]
\centering%
% \vspace{-0.2in}
\includegraphics[width=0.7\linewidth]{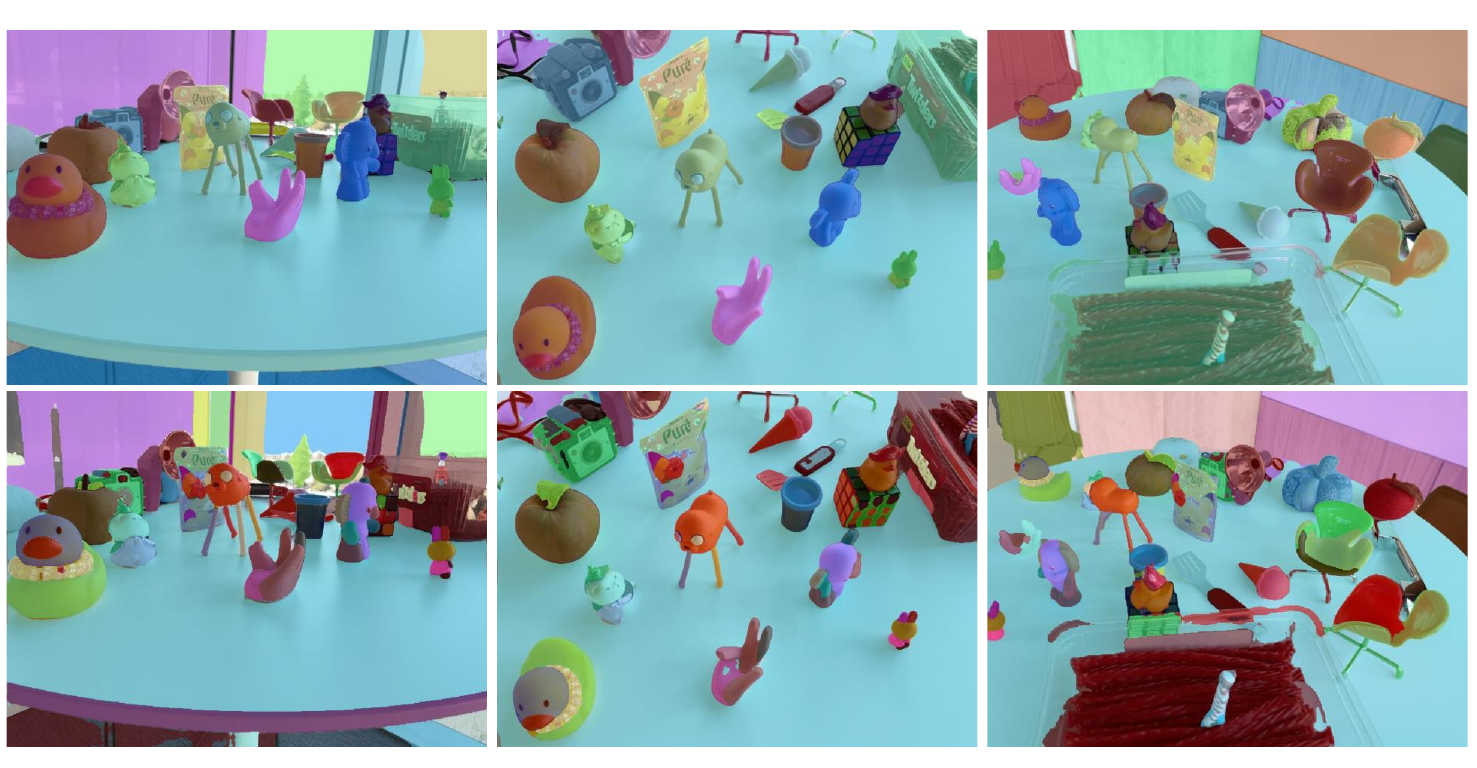}\vspace{-0.15in}%
\caption{Consistent tracking results at multi-granularity masks. The 1st row is in coarse granularity, and the 2nd row is fine-grained.}\vspace{-0.15in}%
\label{fig:granularity}%
\end{figure}

\parsection{Multi-Granularity of Masks}
 As in Fig.~\ref{fig:granularity}, our method can process SAM's anything masks at different granularity levels in the multiple views as follows:
 
\textit{Step \textbf{1}).} Firstly, SAM predicts dense anything mask proposals (including both large-granularity and small-granularity ones) for each frame. 

\textit{Step \textbf{2}).} Then, these masks are scored and sorted based on their mask areas. Masks are ranked in descending (from large to small) or ascending (from small to large) order to pick varying levels of granularity for the label of each pixel.

\textit{Step \textbf{3}).} Furthermore, we filter large-area or small-area masks through their overlapping IoU with a threshold.  

\textit{Step \textbf{4}).} For mask association, masks are temporally propagated in a bi-direction to obtain in-clip consensus. Refer to Sec 3.2 of Deva~\cite{cheng2023tracking} for details.

\parsection{Quantitative Evaluation of Editing} Following Instruct-NeRF2NeRF~\cite{instructnerf2023}, we provide the CLIP Text-Image Direction Similarity evaluation for three of our editing tasks in Table~\ref{exp:editing_clip_score}. Gaussian grouping supports versatile scene editing tasks, including 3D Object Inpainting, 3D Object Style Transfer and 3D Object Removal. 
For each of these editing tasks, Gaussian Grouping outperforms the corresponding SOTA method specifically designed for it.
For inpainting, the descriptions of original and edited scenes are "Lego toy on the table" and "A flat table". For style transfer, the descriptions are "A bear statue" and "Turn the bear into a panda".  For object removal, the descriptions are "A truck on urban street" and "Urban street".

\begin{table}[!t]
\centering
\caption{Quantitative comparison on three Scene Editing Tasks using the CLIP Text-Image Direction Similarity~\cite{instructnerf2023}.}
\vspace{-0.13in}
\resizebox{0.9\linewidth}{!}{
\begin{tabular}{c|c|c|cccc}
\toprule
   Task & Dataset &  Model & {CLIP Text-Image Direction Similarity}  \\  \midrule  
\multirow{2}{*}{3D Object Inpainting} & \multirow{2}{*}{MipNeRF360/Kitchen} & SPIN-NeRF~\cite{mirzaei2023spin} & 0.126 \\
& & Ours &  \textbf{0.153}  \\  \midrule
\multirow{2}{*}{3D Object Style Transfer} & \multirow{2}{*}{Instruct-NeRF2NeRF/Bear} & Instruct-NeRF2NeRF~\cite{instructnerf2023} & 0.171 \\
& & Ours & \textbf{0.178}    \\  \midrule
\multirow{2}{*}{3D Object Removal} & \multirow{2}{*}{Tandt/Truck} & DFFs~\cite{kobayashi2022decomposing} & 0.166  \\
&  & Ours & \textbf{0.183}     \\
 \bottomrule
\end{tabular}}
\label{exp:editing_clip_score}
	\end{table}

\begin{figure}[!t]
\centering%
% \vspace{0.05in}
\includegraphics[width=1.0\linewidth]{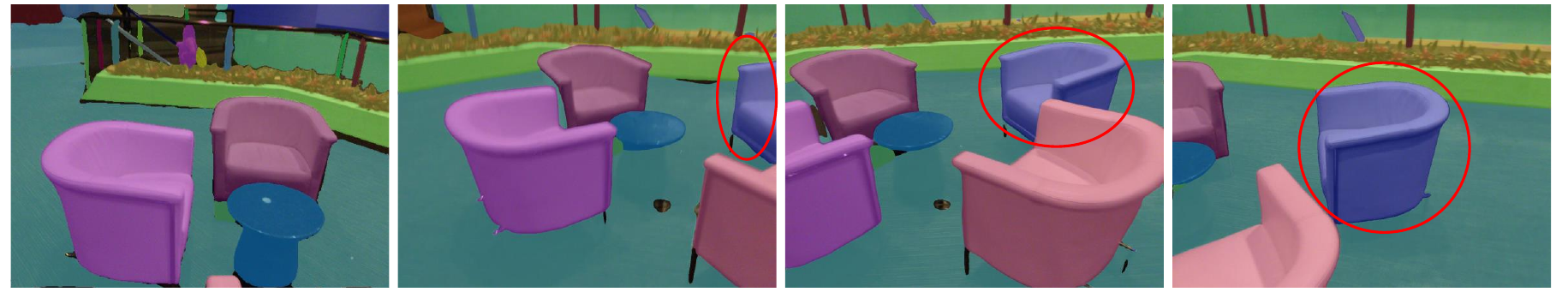}\vspace{-0.15in}%
\caption{Tracking results for a new object in subsequent frames. The \textcolor{red}{red circle} highlights the new chair appearing in the 2nd frame.}\vspace{-0.15in}%%
\label{fig:new_obj}%
\end{figure}

\parsection{Robustness of New Objects} 
 SAM+Tracking can process new objects on subsequent frames. When new high-confident segmentation doesn't match the previous objects and has high confidence, a new instance ID is assigned to it (see the new chair in the red circle in Fig.~\ref{fig:new_obj}).

\begin{figure}[!t]
\centering%
\vspace{-0.1in}
\includegraphics[width=0.75\linewidth]{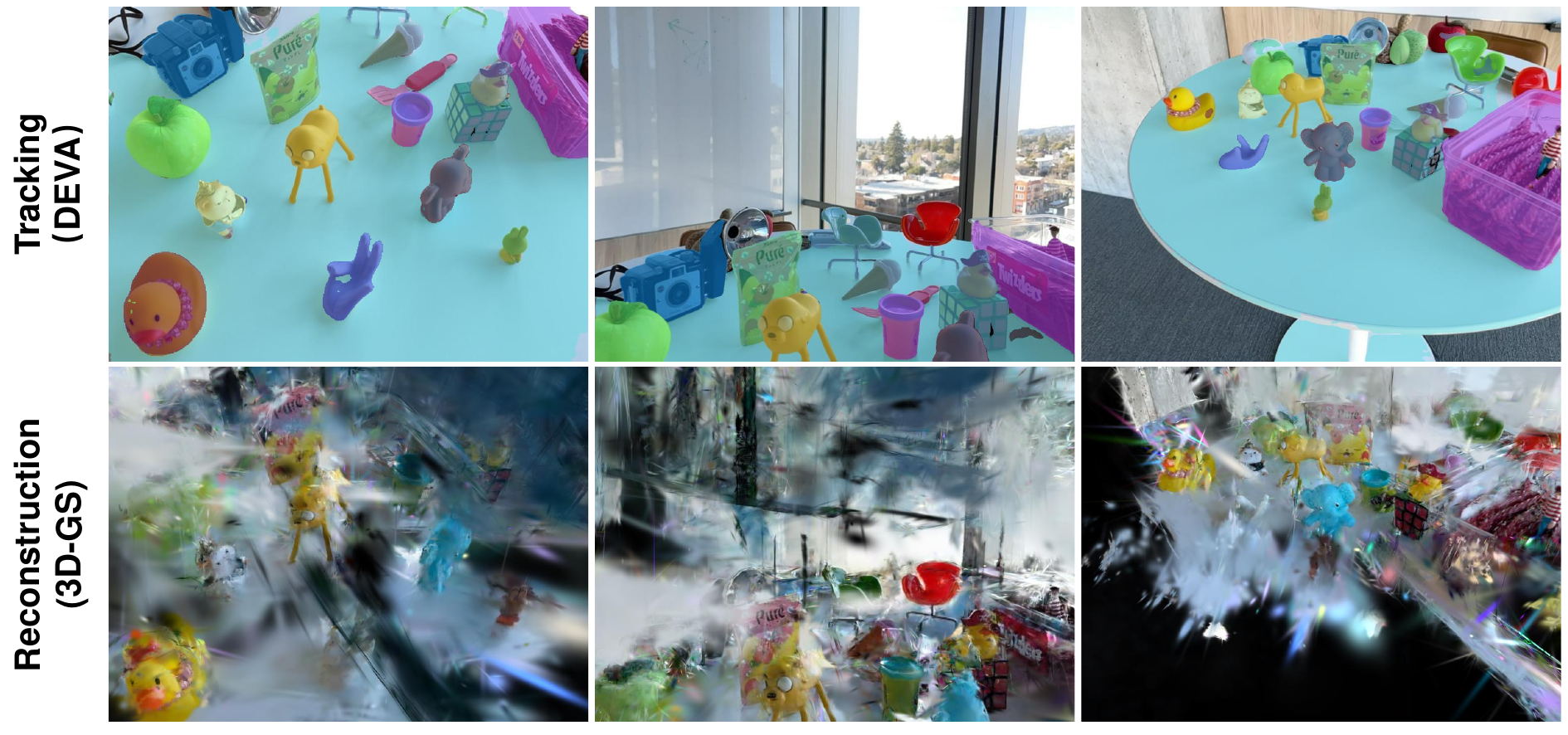}\vspace{-0.15in}%
\caption{Sparse 3-view tracking and reconstruction comparison.}\vspace{-0.2in}%
\label{fig:sparse}%
\end{figure}

\parsection{Sparse View Input} 
For few-shot or sparse-view input, video tracking still obtains a good result for pre-processing, since it is visual feature-based and camera-motion robust. In Figure \ref{fig:sparse}, we use a 3-view input for DEVA pre-processing and still get a decent tracking result. 3D reconstruction of the original Gaussian Splatting is not good with sparse-view input, but it is beyond the scope of this paper. 
It is retained as a component for sparse-view Gaussian Splatting reconstruction, intended for further research in the future.

\subsection{Details on the LERF-Mask Annotation}
\label{sec:dataset}

\parsection{Annotation Pipeline}
To measure the segmentation or fine-grained localization accuracy in the open-wold 3D scene, we construct the LERF-Mask dataset based on the existing LERF-Localization~\cite{lerf2023} evaluation dataset, where we manually annotate three scenes from LERF-Localization with accurate masks instead of using coarse bounding boxes. For each 3D scene, we provide 7.7 text queries with corresponding GT mask labels on average. We use Roboflow~\cite{dwyer2022roboflow} platform for label annotation, and it uses SAM~\cite{kirillov2023segment} as an auxiliary tool for interactive segmentation. Similar to the annotation used in LERF, for each of the 3 scenes, we choose 2-4 novel views for testing and annotating the rendering of novel views.

\begin{figure}[!h]
	\centering
	% \vspace{-0.05in}
	\includegraphics[width=1.0\linewidth]{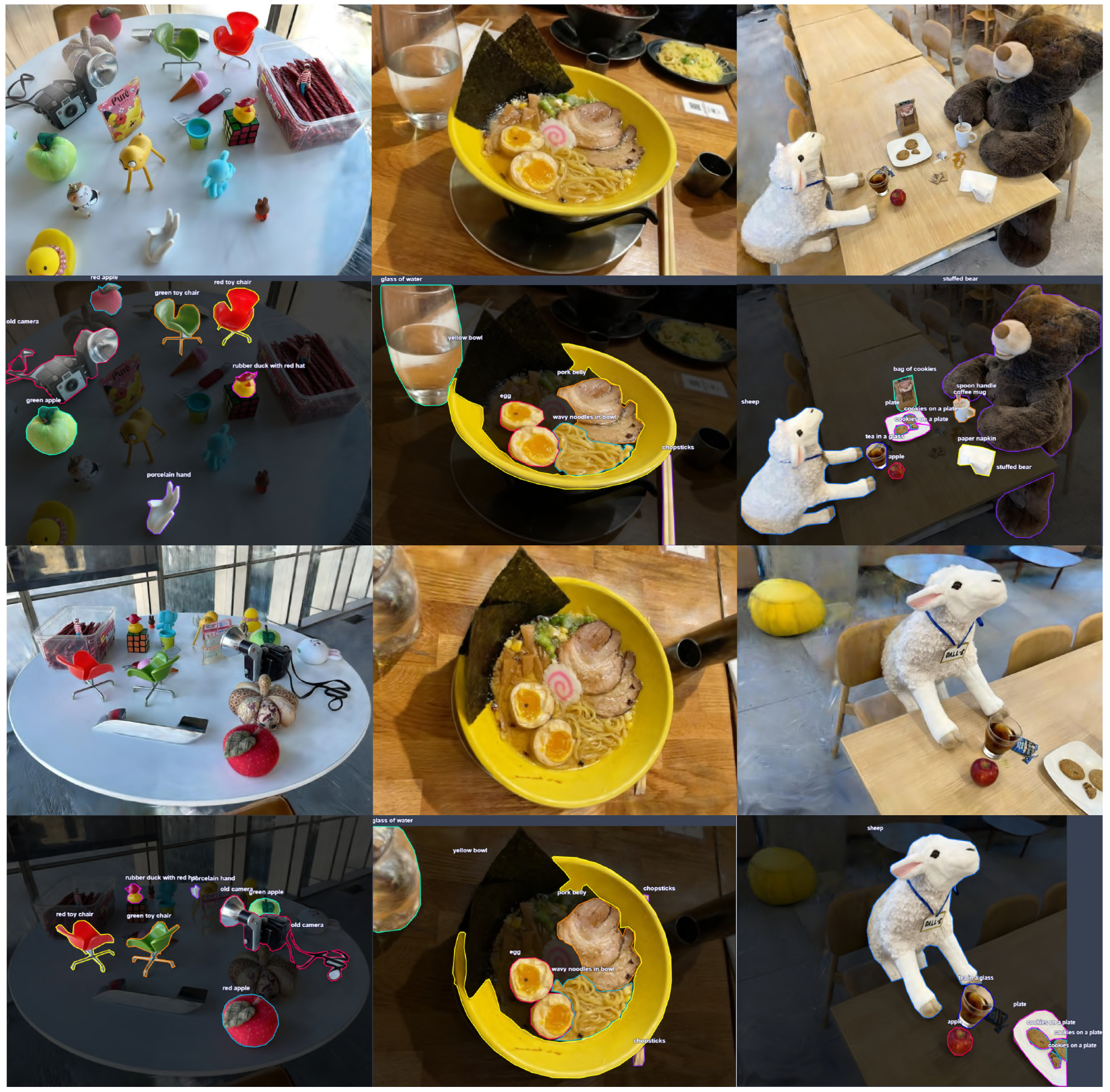}
	\vspace{-0.25in}
	\caption{Annotation visualization of our proposed LERF-Mask dataset. We manually annotate three scenes from LERF-Localization~\cite{lerf2023} with accurate masks instead to replace the coarse bounding boxes in~\cite{instructnerf2023}. The text queries are detailed in Table~\ref{tab:text_labels}.}
	%\label{fig:model}
	\label{fig:annotation}
	\vspace{-0.01in}
\end{figure}

\parsection{Annotation Examples} 
All language prompts used for our LERF-Mask dataset evaluation are listed in Table~\ref{tab:text_labels}, which contains 23 prompts in total. Also, we provide visualization on the mask annotations in Figure~\ref{fig:annotation}.

\begin{table}[h!]
    \centering
    \resizebox{0.7\linewidth}{!}{
    \begin{tabular}{l|cccc}
    \toprule
    Scene & \multicolumn{3}{c}{Text queries} \\
    \toprule
                & green apple & green toy chair & old camera \\
Figurines       & porcelain hand  & red apple & red toy chair \\
                & rubber duck with red hat \\
\midrule
Ramen           & chopsticks & egg & glass of water \\
                & pork belly & wavy noodles in bowl & yellow bowl \\
\midrule
                & apple & bag of cookies & coffee mug \\
Teatime         & cookies on a plate & paper napkin & plate\\
                & sheep & spoon handle & stuffed bear \\
                & tea in a glass  \\
                \bottomrule
    \end{tabular}}
    
    \caption{Prompt labels used during segmentation experiments in our proposed LERF-Mask dataset (23 total).}
    
    \label{tab:text_labels}
\end{table}

\vspace{-0.6in}
\subsection{Local Gaussian Editing: Steps of Object Inpainting \& Style Transfer}
\label{sec:inpaiting_pipe}
\subsubsection{3D Object Inpainting Pipeline} For inpainting, we remove the 3D Gaussians of the selected target by using a Gaussian Grouping model well-trained for 3D reconstruction and segmentation and add new Gaussians for finetuning. 
The steps are as follows:

\textit{Step \textbf{1}).} Train the Gaussian Grouping model with our proposed 2D and 3D Identity Grouping loss.

\textit{Step \textbf{2}).} Select the target object for inpainting. For each Identity Encoding associated with a 3D Gaussian, we acquire its linear layer classification result. Subsequently, we remove those 3D Gaussians that are classified as the label of the selected object. Also, we remove the 3D Gaussians with position inside the convex hull of the object Gaussians.

\textit{Step \textbf{3}).} On the rendering views after the deletion of the object, we detect the ``blurry hole" with Grounding-DINO~\cite{liu2023grounding} as the mask for 2D inpainting and use DEVA~\cite{cheng2023tracking} for association. We use LAMA~\cite{lama} inpainting on each view as the target for finetuning.

\textit{Step \textbf{4}).} After the 3D Gaussians of the target object are deleted, we clone 200K new Gaussians near the deletion region. We freeze the other Gaussians, and only finetune the newly introduced 3D Gaussians. 

\textit{Step \textbf{5}).} During the finetuning, we employ L1 loss only in the outside regions of the object mask, and adopt LPIPS loss inside the bounding box of the object mask. \\

\subsubsection{3D Object Style Transfer Pipeline} For style transfer, we finetune the 3D Gaussians belonging to the corresponding target by using a Gaussian Grouping model well-trained for 3D reconstruction and segmentation. 
The steps are as follows:

\textit{Step \textbf{1}).} Train our Gaussian Grouping model with our proposed 2D and 3D Identity Grouping loss.

\textit{Step \textbf{2}).} Select the target object for style transfer. For each Identity Encoding associated with a 3D Gaussian, we acquire its linear layer classification result. Subsequently, we only finetune those 3D Gaussians that are classified as the label of the selected object. 
Also, we finetune the 3D Gaussians with position inside the convex hull of the selected object. The Gaussians irrelevant to the editing target are frozen.
During finetuning, we freeze the 3D position of Gaussains and make other Gaussian parameters (color, variance, opacity, etc.) trainable.

\textit{Step \textbf{3}).} During the finetuning process, we dynamically update the target images using an image-level style transfer model that has been pre-trained. Specifically, we employ InstructPix2Pix\cite{brooks2022instructpix2pix}, introducing a noise input composed of the rendered view combined with random noise. This approach involves conditioning the diffusion model on a ground truth image to enhance accuracy and consistency.

\textit{Step \textbf{4}).} To preserve the spatial details of the background regions, rendering losses are exclusively performed within the mask of the style transfer target. On the 2D rendered view, we employ L1 loss inside the object mask and LPIPS loss within the bounding box that encloses the object mask.

\subsection{More Implementation Details}
\label{sec:supp_details}

\subsubsection{More implementation details} 
We implement Gaussian Grouping based on Gaussian Splatting~\cite{kerbl20233d}. We add a 16-dimension identity encoding as a feature of each Gaussian, and implement forward and backward cuda rasterization similar to the direct current of Spherical Harmonics. The 3D Identity Encoding has a shape of $N*1*16$, where $N$ is the number of Gaussians. We set the degree of Spherical Harmonics to zero since instance identity does not change across views. 
The rendered 2D Identity Encoding has a shape of $16*H*W$. 
3D Identity Encoding and 2D Identity Encoding share the same identity classification linear layer with $(16,256)$ input and output channels.

During training, we set $\lambda_\text{2d} = 1.0$ and $\lambda_\text{3d} = 2.0$.  We use the Adam optimizer for both Gaussians and the linear layer, with a learning rate of 0.0025 for identity encoding and 0.0005 for the linear layer. For 3D regularization loss, we set the nearest neighboring number to $k=5$, and the sampling points number to $m=1000$. To improve efficiency and avoid calculating loss at boundary points, we downsample the point cloud to 300K to calculate the loss. 3D regularization loss only affects the segmentation of identity encoding and does not affect the density of the Gaussians. We use the same adaptive density control as Gaussian Splatting. All datasets are trained for 30K iterations on one A100 GPU.

\subsection{Limitation}
Our Gaussian Grouping segments ``anything masks'' with the assistance of SAM. But the ``anything mask labels" by original SAM~\cite{kirillov2023segment} have no direct semantic language information. We adopt the Grounding-DINO\cite{liu2023grounding} for open vocabulary segmentation to pick the 2D object, and match our anything masks rendering. When some language prompts are very complicated, the Grounding-DINO can not acquire the correct mask from the input text prompt and will give a wrong mask prediction. In this case, even if we provide the correct mask in anything mask rendering, we do not obtain explicit category information. 
Also, the zero-shot 2D association accuracy of DEVA~\cite{cheng2023tracking} will also limit the open-world 3D segmentation performance of Gaussian Grouping.
This can be solved by further improvement of the vision language detection model and better association schemes in the future.

% \clearpage\mbox{}Page \thepage\ of the manuscript.
% \clearpage\mbox{}Page \thepage\ of the manuscript.
% \clearpage\mbox{}Page \thepage\ of the manuscript.
% \clearpage\mbox{}Page \thepage\ of the manuscript.
% \clearpage\mbox{}Page \thepage\ of the manuscript. This is the last page.
% Now we have reached the maximum length of an ECCV \ECCVyear{} submission (excluding references).
% References should start immediately after the main text, but can continue past p.\ 14 if needed.
\clearpage  % TODO REVIEW/FINAL: This \clearpage needs to be removed from both review and camera-ready versions.

\bibliographystyle{splncs04}
\bibliography{main}
\end{document}